\documentclass[10pt]{wlscirep}
\usepackage{amsmath}
\usepackage{amsfonts}
\usepackage{amssymb}

\usepackage{amsthm}
\usepackage{subcaption}
\usepackage{cite}
 \usepackage{multirow} 
\usepackage{algorithmic}
\usepackage{url}
\usepackage{booktabs}
\usepackage{tikz,pgfplots}
\pgfplotsset{compat=newest}
\usetikzlibrary{external}
\tikzsetexternalprefix{tikzfigures/}
\usepgfplotslibrary{fillbetween}
\usepackage{url}
\usepackage{amsmath}
\usepackage{amsfonts}
\usepackage{amssymb}
\usepackage{float} 
\usepackage[ruled, vlined, nofillcomment]{algorithm2e}
\usepackage{algorithmic}
\usepackage{booktabs}
\usepackage{float}

\SetCommentSty{mycommfont}
\SetAlgoCaptionSeparator{:}

\usepackage{graphicx}
\usepackage{tikz}
\usepackage{framed}
\newcommand{\Transp}{\mathsf{T}}

\newcommand{\myindent}[1]{\parbox[t]{\linewidth}{\quad \quad}}

\newlength\figureheight 
\newlength\figurewidth 
\title{A Tightly Coupled IMU-Based Motion Capture Approach for Estimating Multibody Kinematics and Kinetics}

\author[1,*]{Hassan Osman}
\author[2]{Daan de Kanter}
\author[1,2]{Jelle Boelens}
\author[2]{Manon Kok}
\author[1]{Ajay Seth}
\affil[1]{Department of Biomechanical Engineering, Delft University of Technology, the Netherlands}
\affil[2]{Delft Center for Systems and Control, Delft University of Technology, the Netherlands}

\affil[*]{h.osman@tudelft.nl}

\begin{abstract}
Inertial Measurement Units (IMUs) enable portable, multibody motion capture (MoCap) in diverse environments beyond the laboratory, making them a practical choice for diagnosing mobility disorders and supporting rehabilitation in clinical or home settings. However, challenges associated with IMU measurements, including magnetic distortions and drift errors, complicate their broader use for MoCap. In this work, we propose a tightly coupled motion capture approach that directly integrates IMU measurements with multibody dynamic models via an Iterated Extended Kalman Filter (IEKF) to simultaneously estimate the system’s kinematics and kinetics. By enforcing kinematic and kinetic properties and utilizing only accelerometer and gyroscope data, our method improves IMU based state estimation accuracy.
Our approach is designed to allow for incorporating additional sensor data, such as optical MoCap measurements and joint torque readings, to further enhance estimation accuracy. We validated our approach using highly accurate ground truth data from a 3 Degree of Freedom (DoF) pendulum and a 6 DoF Kuka robot. We demonstrate a maximum Root Mean Square Difference (RMSD) in the pendulum's computed joint angles of 3.75° compared to optical MoCap Inverse Kinematics (IK), which serves as the gold standard in the absence of internal encoders. For the Kuka robot, we observe a maximum joint angles RMSD of 3.24° compared to the Kuka's internal encoders while the maximum joint angles RMSD of the optical MoCap IK compared to the encoders was 1.16°. Additionally, we report a maximum joint torque RMSD of 2 Nm in the pendulum compared to optical MoCap Inverse Dynamics (ID), and 3.73 Nm in the Kuka robot relative to its internal torque sensors.
\end{abstract}
\begin{document}

\flushbottom
\maketitle
%
%
\thispagestyle{empty}

\section*{Introduction}

Measurements from Motion Capture (MoCap) systems are used to capture and analyze multibody movements \cite{menache2010understanding,MOESLUND200690}. This multibody movement analysis is essential for a wide range of applications, including human rehabilitation, sports, film animation, and video games \cite{Suo24,LIAO2020,Ben2009}. To achieve this, MoCap systems provide measurements from a collection of sensors, from which multibody kinematics and kinetics can be calculated. These measurements can take the form of 3D body marker positions in marker-based MoCap, 3D body orientations in Inertial Measurement Unit (IMU)-based MoCap, or joint center positions in markerless optical MoCap \cite{Salisu23}.

Inertial sensors allow multibody MoCap in various environments and settings, unlike optical marker-based systems, which are typically restricted to controlled laboratory conditions \cite{Das2023,Liu2020}. This flexibility makes IMU-based MoCap particularly valuable for applications such as patient rehabilitation, where portability and accessibility are important \cite{LIAO2020}. However, due to accuracy and reliability limitations, IMU-based MoCap has not been widely used in clinical applications \cite{Fang23,Cho2018}. Our aim is to develop a tightly coupled IMU-based MoCap approach that leverages knowledge of the multibody system's properties alongside IMU measurements to estimate kinematics and kinetics. Our approach also integrates complementary measurements, such as optical MoCap data and zero-torque updates, to enhance estimation accuracy.

IMUs capture linear acceleration and angular velocity using accelerometers and gyroscopes \cite{Kok2017}. Some IMUs also incorporate a magnetometer to measure magnetic field intensity. Using these measurements, the orientation of the IMU can be computed \cite{Kok2017}. Inaccuracies in IMU-derived orientations arise from multiple sources: magnetic distortions (especially indoors), gyroscope bias and noise, and the difficulty of separating inertial accelerations from gravitational acceleration during motion. Consequently, orientation estimates contain errors that propagate through subsequent analyses \cite{W.H.K.09,Kok2017,vanDijk2021,Kok2019}.

OpenSense is an open-source algorithm that utilizes IMU orientations together with the system’s kinematic model to estimate multibody kinematics via inverse kinematics (IK) \cite{AlBorno2022}. In this method, virtual IMUs are placed on the model’s body segments with the same initial orientations as the real IMUs. Joint angles are then computed by minimizing the error between virtual and experimental IMU orientations during motion. However, accurate IMU orientation estimates are critical for reliable kinematics with OpenSense.

Joint kinematics can be coupled with IMU measurements by enforcing that the acceleration at the joint center shared by two segments is identical for the IMUs mounted on each segment \cite{Weygers2020}. This coupling leverages the system’s kinematic constraints alongside sensor data to improve motion estimates. Since kinetics are not directly estimated, a common approach is to use the estimated kinematics to compute kinetics via inverse dynamics (ID) in a second step \cite{winter1990biomechanics,Kuo98,RIEMER20081503,Faber2018}. However, this two-step process allows kinematic errors to propagate into the kinetics estimates \cite{Kuo98,Ojeda2016}. Instead, we our pipeline jointly estimates both kinematics and kinetics by integrating sensor measurements and the system’s dynamic model.

Incorporating a dynamic model has been previously found to improve the accuracy of the estimated kinematics and kinetics during MoCap. One way to achieve this is by formulating an optimal control problem that computes movement and control trajectories using dynamic models and raw inertial data (accelerometer and gyroscope measurements) \cite{Dorschky2019,Haraguchi2024,Nitschke2024}. However, these approaches either employed a planar model rather than a full 3D model or relied on simulated inertial measurements generated from marker-based MoCap data. Our approach employs an Iterated Extended Kalman Filter (IEKF) algorithm to integrate an accurate 3D dynamic model with actual IMU measurements. This framework can be extended to incorporate different sensor modalities and measurements, enabling more accurate kinematics and kinetics estimates. Our pipeline was tested using real sensor measurements on two experimental setups-a 6 DoF Kuka robot and a 3 DoF pendulum—each of which provided accurate ground truth data.
\section{Methodology}
\label{methodlgy}
\begin{figure}
        \centering
        \includegraphics[width=.8\linewidth]{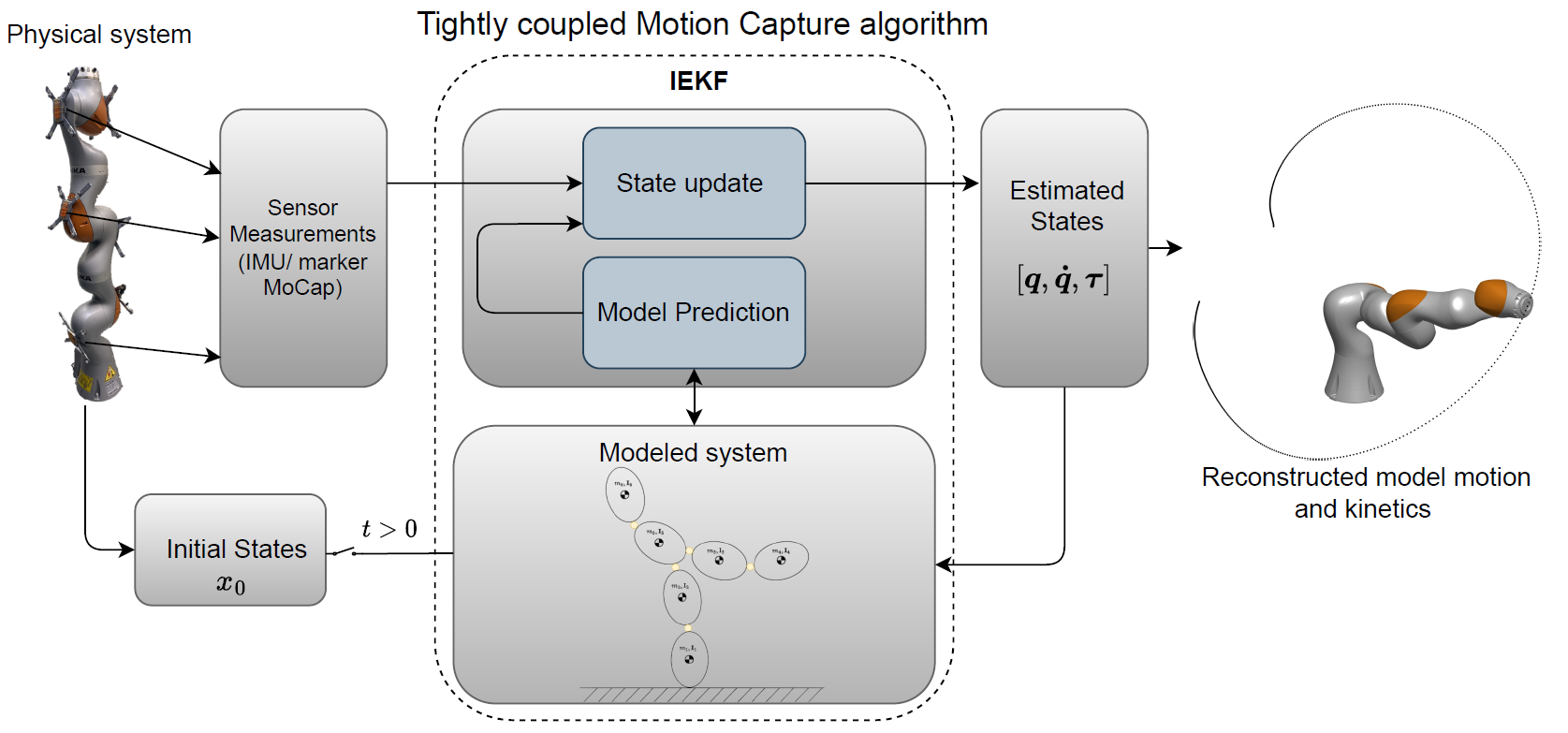}
    \caption{Overview of the tightly coupled sensor and model dynamics approach. Given the initial states, IEKF couples sensor measurements and the multibody system model to estimate generalized joint angles, velocities, and torques.}
    \label{fig:Pipeline1}
\end{figure}

The general outline of the algorithm shown in Figure~\ref{fig:Pipeline1} illustrates the motion estimation process where the sensors measurements are coupled with the system multibody model through an Iterative Extended Kalman Filter (IEKF). The IEKF algorithm utilizes the system multibody model along with the inital or the previous states to predict the current states then both the senor measurements and the measurement model are used to iteratively update this prediction to estimate the current states. In this work we levarage OpenSim which is an open source software used to model and simulate musculoskeletal systems to  model and evaluate the multibody system's equations of motion and kinematic jacobians during the estimation process~\cite{Delp2007,Seth2018OpenSim,Seth2011,Vivian2016}. We also use OpenSim to derive the measurement model by integrating the sensors measurements with the multibody system. We demonstrate such integration by discussing the general multibody dynamics model, the IMU, marker MoCap and zero torque updates measurement models and the IEKF approach used to couple between them to estimate the motion's kinematics and kinetics simultaneously.    

\begin{figure}
\centering
\includegraphics[width=0.65\linewidth]{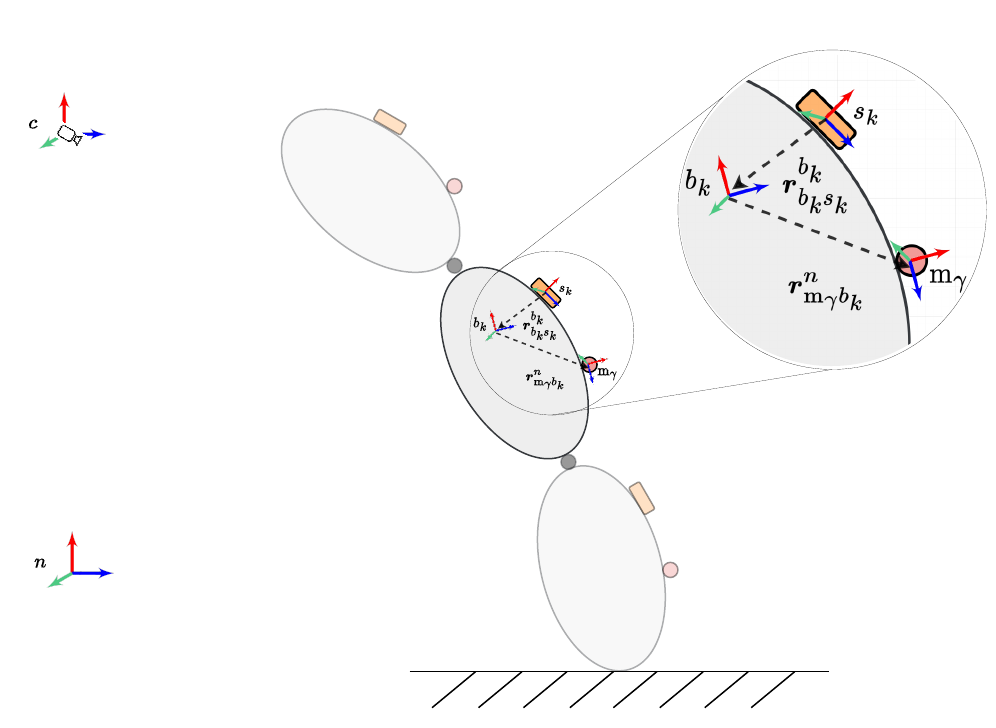}

\caption{Displays the definitions of the navigation frame $n$, the optical MoCap frame $c$, the body frame $b_k$, the IMU frame $s_k$, and the marker frame $m_\gamma$. It also includes the translation vectors $\boldsymbol{r}^{b_k}_{b_k s_k}$ and $\boldsymbol{r}^n_{m_{\gamma} b_k}$}

\label{fig:BodiesAndJoints}
\end{figure}

\label{sec:modeling}
The motion is estimated for a dynamic chain with $N_J$ joints, $N_B$ body segments and $N_D$ generalized coordinates represented as joint angles at each time instance $t$. The estimated motion states are expressed as the generalized joint angles $\textbf{\textit{q}} = \begin{pmatrix} q_{1},q_{2}, & \cdots &, q_{N_D} \end{pmatrix}^\Transp$, the joints angular speeds $\dot{\textbf{\textit{q}}}
 = \begin{pmatrix} \dot{q}_{1},\dot{q}_{2}, & \cdots &, \dot{q}_{N_D} \end{pmatrix}^\Transp$ and the generalized forces corresponding to the joint torques $\boldsymbol{\tau} = \begin{pmatrix}\tau_{1},\tau_{2}, & \cdots &,\tau_{N_D}\end{pmatrix}^\Transp$. Hence the state vector $x$ that we are interested in estimating is given by 
\begin{equation}
    \boldsymbol{x^\Transp} = \begin{pmatrix} 
    \boldsymbol{q^\Transp} & \boldsymbol{\dot{q}^\Transp} & \boldsymbol{\tau^\Transp}
    \end{pmatrix},
    \label{eq:state}
\end{equation}
 where $\boldsymbol{q} \in \mathbb{R}^{N_D}$, $\boldsymbol{\dot{q}} \in \mathbb{R}^{N_D}$ and  $\boldsymbol{\tau}\in \mathbb{R}^{N_D}$.
Figure~\ref{fig:BodiesAndJoints} illustrates the relation between the IMUs and the markers with the body segments where we assume that the sensors are rigidly attached to the body.

\subsection{Multibody dynamic model}
\label{sec:dynModel}

OpenSim framework was used to model the multibody system and capture the dynamics that describes the states in~\eqref{eq:state}. The system dynamics are presented as 
\begin{equation}
\boldsymbol{\tau} = \bar{\textbf{M}}(\boldsymbol{q})\ddot{\boldsymbol{q}} + \underbrace{\textbf{C}(\boldsymbol{q}, \dot{\boldsymbol{q}}) + \textbf{G}(\boldsymbol{q})}_{\boldsymbol{\tau}^I(q,\dot{q})},
\end{equation}
for which $\bar{\textbf{M}}(\boldsymbol{q})$ is a symmetric, positive definite mass matrix, $\textbf{C}(\boldsymbol{q}, \dot{\boldsymbol{q}})$ is a function that groups all the velocity terms and $\textbf{G}(\boldsymbol{q})$ is a function that represents the gravity force component .$\tau^I(q,\dot{q})$ is a function representing the torque that results from gravity, Coriolis force and centrifugal force. Hence, the joints angular accelerations can be expressed as,
\begin{equation}
\boldsymbol{\ddot{q}} = \bar{\textbf{M}}(\boldsymbol{q})^{-1}(\boldsymbol{\tau}-\boldsymbol{\tau}^I(\boldsymbol{q},\boldsymbol{\dot{q}})),
\label{qdotdot}
\end{equation}
The system dynamic model is expressed to include process noise, which accounts for uncertainties and disturbances in the system as, 
\begin{equation}
\dot{\mathbf{x}} = g(\boldsymbol{q}, \dot{\boldsymbol{q}}, \boldsymbol{\tau}) + {e}_{p},
\label{eq:stateSpaceModel}
\end{equation}
where ${e}_{p} \sim \mathcal{N}(0, \mathbf{Q})$ is the process noise and $\mathbf{Q} \in \mathbb{R}^{3N_D \times 3N_D}$ is the process noise covariance matrix. 
\subsection{General measurement model}
\label{sec:GmeasModel} 
The process of estimating the system states ~\eqref{eq:state} via directly coupling the sensor measurements with the system dynamics requires formulating a measurement model. This model describes the system states in terms of the measurements~\cite{Kok2017,Kok2014,Escalona21,Lugris2024}.
Hence, the system measurement model could be expressed as, 
\begin{equation}
\boldsymbol{y}=h(\boldsymbol{x})+e_{meas}
\end{equation}
where $e_{meas}\mathcal\sim{N}(0,\bold{R})$ is the measurement error and $h(x)\in \mathbb{R}^{n}$ is a function that maps the states to the $n$ sensor measurements. Using the states in~\eqref{eq:state} the measurement model is expressed as
\begin{equation}
\boldsymbol{y} = \begin{pmatrix}
h_1(\boldsymbol{q}, \dot{\boldsymbol{q}}, \boldsymbol{\tau}) \\
h_2(\boldsymbol{q}, \dot{\boldsymbol{q}}, \boldsymbol{\tau}) \\
\vdots \\
h_n(\boldsymbol{q}, \dot{\boldsymbol{q}}, \boldsymbol{\tau})
\end{pmatrix}+e_{meas}.
\label{generalmeamodel}
\end{equation}
\subsection{IMU sensor measurement model}
\label{sec:measModel} 
The total number of IMUs fitted on the system's body segments is $N_I$. Each IMU $s_k$, $k=1,\hdots,N_I$ has both a gyroscope that measures angular velocities $\boldsymbol{{\omega}}^{s_k}_{s_k}$ and an accelerometer that measures linear accelerations ${\dot{{\boldsymbol{v}}}}^{s_k}_{s_k}=\boldsymbol{{a}}^{s_k}_{s_k}-\boldsymbol{{g}}$ where the subscript $s_k$ donates the measurement frame and the superscript $s_k$ donates the frame at which the measurement is expressed in. This notation will be used thoughout the rest of this paper. 
The frame $s_k$ axes are assumed to be aligned with the sensor’s axes, with its origin located at the center of the accelerometer triad. The measurement model for the gyroscope angular velocity is given as 
\begin{equation}
    \boldsymbol{y}_{\omega_{k}}
    = \boldsymbol{\omega}_{\text{s}_{k}}^{\mathrm{s}_k} + {e}_{\omega}^{\mathrm{s}_k}, 
    \label{eq:gyrModel}%
\end{equation} 
where $\boldsymbol{y}_{\omega}\in\mathbb{R}^{3\times1}$ is the measured angular velocities and  $e_{\omega}^{\mathrm{s}_k} \sim \mathcal{N}(0,\Sigma_{\omega,k})$ is the measurement noise. Measured angular velocities can be expressed in terms of the body segments angular velocities which are dependent on the system current generalized coordinates and velocities~\cite{Featherstone2008}  as,
\begin{equation}
    \boldsymbol{y}_{\omega_k} = \boldsymbol{R}^{\text{s}\text{b}_k} \boldsymbol{R}^{\text{b}_k \text{n}} (\boldsymbol{q}) \boldsymbol{\omega}_{\text{b}_k}^{\mathrm{n}} (\boldsymbol{q},\boldsymbol{\dot{q}}) + e_{\omega}^{\mathrm{s}_k}, \quad 
    \label{eq:gyrModel-state}
\end{equation}
here two rotation matrices are introduced. The constant rotation matrix $\boldsymbol{R}^{\text{s} \text{b}_k}$ between the body frame $b_k$ and the sensor frame $s_k$ is assumed to be known, $\boldsymbol{R}^{\text{b}_k \text{n}}$ is the rotation matrix that describes the orientation transformation from the navigation frame $n$ to the body segment frame $b_k$ for different generalized joint angles.  

The IMU's accelerometer is capable of measuring the linear accelerations that it undergoes including the acceleration due to gravity $\boldsymbol{g}^n$.  The measured acceleration is expressed as       

\begin{equation}
    \boldsymbol{y}_{{\dot{{{v}}}}} =
    \boldsymbol{R}^{\mathrm{s}_k\mathrm{n}} (q)(\boldsymbol{a}_{\mathrm{s_k}}^\mathrm{n}(\boldsymbol{q},\boldsymbol{\dot{q}},\boldsymbol{\ddot{q}})-\boldsymbol{g}^\mathrm{n})+\boldsymbol{e}_{\dot{{{v}}}}^{\mathrm{s}_k},
    \label{eq:accModel}
\end{equation}
here $\boldsymbol{y}_{{\dot{{v}}}}\in\mathbb{R}^{3\times1}$ is the measured linear accelerations. The rotation matrix $\boldsymbol{R}^{\text{s}_k \text{n}}$ is used due to the fact that although the gravitational acceleration is constant in the navigation frame $n$ while its not constant in the senor frame since, the sensor orientation is changing due to the motion. The accelerometer measurement noise is defined as $e_{{\dot{{{v}}}}}^{\mathrm{s}_k} \sim \mathcal{N}(0,\Sigma_{{\dot{{{v}}}},k})$. 
 The measured accelerations are dependent on the current generalized joint angles, angular velocities and angular accelerations. Therefore, using~\eqref{qdotdot} and~\eqref{eq:state} the measured accelerations can be expressed in terms of the system states as, 
\begin{equation}
    \boldsymbol{y}_{{\dot{{{v}}}}} =
    \boldsymbol{R}^{\mathrm{s}\mathrm{b}_k} \boldsymbol{R}^{\mathrm{b}_k\mathrm{n}} (\boldsymbol{q}) (\boldsymbol{a}_{\mathrm{s_k}}^\mathrm{n}(\boldsymbol{x}) - \boldsymbol{g}^\mathrm{n}) + e_{\text{a}}^{\mathrm{s}_k}.
    \label{eq:accModel-state}
\end{equation}
The Acceleration $\boldsymbol{a}_{\mathrm{s}_k}^\mathrm{n}(\boldsymbol{x})$ is defined in terms of the body's angular velocity, angular acceleration and linear acceleration as   
\begin{align}
    \boldsymbol{a}_{\mathrm{s}_k}^\mathrm{n} (\boldsymbol{x}) &= \boldsymbol{a}_{\mathrm{b}_k}^\mathrm{n} (\boldsymbol{x}) + [\boldsymbol{\alpha}_{\text{b}_k}^\mathrm{n} (\boldsymbol{x}) \times] \left( \boldsymbol{R}^{\mathrm{n} \mathrm{b}_k} (\boldsymbol{q}) \boldsymbol{r}_{b_ks_k}^{\mathrm{b}_k} \right) + \nonumber \\ 
    &\qquad [\boldsymbol{\omega}_{\text{b}_k}^\mathrm{n} (\boldsymbol{q},\boldsymbol{\dot{q}}) \times]^2 \left( \boldsymbol{R}^{\mathrm{n} \mathrm{b}_k} (\boldsymbol{q}) \boldsymbol{r}_{b_ks_k}^{\mathrm{b}_k} \right).
\end{align}
Where $\boldsymbol{a}_{\text{b}_k}^{\mathrm{n}}$,$\boldsymbol{\alpha}_{\text{b}_k}^{\mathrm{n}}$ and $\boldsymbol{\omega}_{\text{b}_k}^{\mathrm{n}}$ are respectively the body segement's linear acceleration, angular acceleration and  angular velocity and  $\boldsymbol{r}_{b_ks_k}^{\mathrm{b}_k}$ is a translation vector expressed in the body frame $b_k$  that maps the position of the IMU frame $s_k$ to the body frame $b_k$ shown in  figure~\ref{fig:BodiesAndJoints}. 
\subsection{Maker based and zero torque update measurement models}
\label{addmeas}
    Marker-based MoCap systems provide marker position information with respect to the system camera frame $c$. Each body segment can have any number of markers attached for a total number of markers $N_M$. We donate each individual marker as $m_{\gamma}$ where  $\gamma=1,\hdots,N_M$. The marker position with respect to the camera frame can then be defined as

\begin{equation}
     \boldsymbol{y}_{\text{p},}=\boldsymbol{P}^{\mathrm{c}}_{m_{\gamma}}+e_{\text{p}}^{\mathrm{m}_{\gamma}}, 
\end{equation}
where $ \boldsymbol{y}_{\text{p}}\in\mathbb{R}^{3\times1}$ is the measured marker position and $e_{\text{p}}^{\mathrm{m}_{\gamma}}\sim \mathcal{N}(0,\Sigma_{\text{p},{\gamma}})$ is the position measurement noise. Marker position measurements can then be written in terms of the body position as  
\begin{equation}
\boldsymbol{y}_{\text{p}} = \boldsymbol{R}^{\mathrm{cb_{k}}}(\boldsymbol{q})\boldsymbol{R}^{\mathrm{b_{k}n}}(\boldsymbol{q})(\boldsymbol{P}^\mathrm{n}_{b_{k}}+\boldsymbol{r}^{\mathrm{n}}_{m_{\gamma}b_k}(\boldsymbol{q}))+e_{\text{p}}^{\mathrm{m}_{\gamma}}.
\end{equation}
The translation vector $\boldsymbol{r}^{\mathrm{n}}_{m_{\gamma}b_k}\in \mathbb{R}^{3 \times 1}$ is a vector that maps the position of the body segment frame $b_k$  to the marker frame $m_\gamma$ with respect to the navigation frame $n$ shown in figure~\ref{fig:BodiesAndJoints}. This vector and the two matrices $\boldsymbol{R}^{\mathrm{b_{k}n}}\in \mathbb{R}^{3 \times 3}$ and $\boldsymbol{R}^{\mathrm{cb_{k}}}\in \mathbb{R}^{3 \times 3}$ are dependent on the generalized joint angles since the bodies and markers frames change orientations with respect to the camera frame $c$ and the navigation frame $n$ during motion.  

To incorporate zero torque updates to the pipeline, we model virtual joint torque sensors $\phi_j$, $j=1,\hdots,N_D$ that assume joint torque measurements of zero. This measurements are expressed as, 
\begin{equation}
    \boldsymbol{y}_{\tau}=0+e_{\tau}^{\phi_j},
\end{equation}
where ${y}_{\tau}\in\mathbb{R}^{3\times1}$ is the joint zero torque virtual measurement and $e_{\tau,t}^{\phi_j}\sim \mathcal{N}(0,\Sigma_{\tau,\text{j}})$ is the virtual torque measurement noise. 
\subsection{Tightly Coupled Kinematics and Kinetics estimation algorithm}
\label{sec:estimation}
\begin{figure}
    \centering
    \includegraphics[width=.75\linewidth]{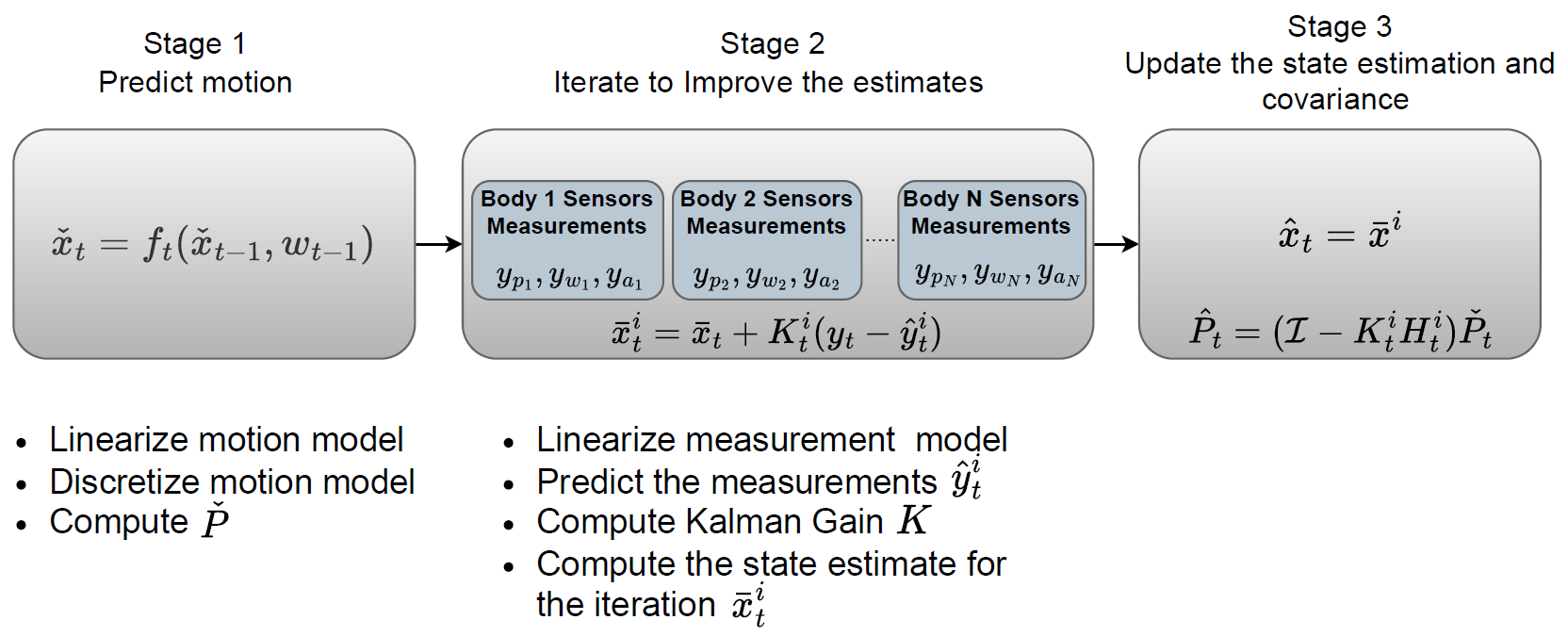}
    \caption{IEKF process: (1) Predict the states using the prior states and the OpenSim model, (2) Refine the state predictions by iteratively linearizing the measurement model either from the IMUs only or the combination of IMUs and other sensors to account for non-linearities then compute the Kalman gain $\boldsymbol{K}$, (3) Update the state estimation and covariance.}
    \label{fig:Pipeline2}
\end{figure}

Our frame-work integrates the dynamic model described in~\ref{sec:dynModel} with the measurement models described in~\ref{sec:GmeasModel}, \ref{sec:measModel} and~\ref{addmeas} using an IEKF pipeline, as shown in Figure~\ref{fig:Pipeline2}. The kinematic and dynamic properties of the system are defined using OpenSim~\cite{seth2010minimal,SHERMAN2011241}. By modeling the sensors within this framework and incorporating them into the system dynamic model, we couple the sensor measurements to the states of the system.  

The dynamic model is discretized to predict the future state at time $t+1$ by numerically integrating the equations of motion using OpenSim's Runge-Kutta-Merson integrator~\cite{SHERMAN2011241}, resulting in an update equation of the form, 
\begin{equation}
\boldsymbol{x}_{t+1} = \boldsymbol{x}_t + \Delta t \cdot g(\boldsymbol{x}_t).
\label{eq:dynModel-discrete}
\end{equation}
The algorithm measurement model that includes all the previously discussed sensors measurement models can then be expressed as
\begin{equation}
\boldsymbol{y} = \begin{pmatrix} 
    \boldsymbol{y}_{\omega,\text{s}_1} \\ 
    \boldsymbol{y}_{{\dot{{{v}}}},\text{s}_1} \\ 
    \boldsymbol{y}_{\text{p},\text{m}_1} \\ 
    \boldsymbol{y}_{\tau,\phi_1} \\ 
    \vdots \\ 
    \boldsymbol{y}_{\omega,\text{s}_{N_I}} \\ 
    \boldsymbol{y}_{{\dot{{{v}}}},\text{s}_{N_I}} \\ 
    \boldsymbol{y}_{\text{p},\text{m}_{N_M}} \\ 
    \boldsymbol{y}_{\text{p},\phi_{N_D}} 
\end{pmatrix}
+ e_{\text{meas}},
\label{meas1}
\end{equation}
where $e_{\text{meas}}\mathcal\sim{N}(0,\bold{R})$ is total measurement error and $\bold{R}\in\mathbb{R}^{3(2N_I+N_M+N_D)\times3(2N_I+N_M+N_D)}$ is the measurement error covariance.
The set of functions in~\eqref{generalmeamodel} that maps the states~\eqref{eq:state} to the sensor measurements using the sensors measurement models and the dynamic model discussed in~\ref{sec:modeling} is hence expressed as 
\begin{equation}
    \boldsymbol{y} = \begin{pmatrix} h_{\omega,1}(\boldsymbol{x}) \\ h_{{\dot{{{v}}}},1}(\boldsymbol{x})\\ h_{\text{p},1}(\boldsymbol{x})\\ h_{\tau,1}(\boldsymbol{x})\\ \vdots \\ h_{\omega,N_I}(\boldsymbol{x}) \\ h_{{\dot{{{v}}}},N_I}(\boldsymbol{x})\\ h_{\text{p},N_M}(\boldsymbol{x})\\h_{\tau,N_D}(\boldsymbol{x}) \end{pmatrix}+ e_{\text{meas}}.
\end{equation}

The process of estimating the system states $\hat{\boldsymbol{x}}_t$ and their covariance $\hat{\boldsymbol{P}}_t$ using an IEKF is described in Alg.~\ref{IEKF_OS_Algorithm}.
The first step is the prediction in which the state and covariance are predicted as $\check{\boldsymbol{x}}_t$ and $\check{\boldsymbol{P}}_t$ respectively. In this step the Jacobian $\boldsymbol{G}=\tfrac{\partial g(\dot{\boldsymbol{x}}_t)}{\partial{\boldsymbol{x}}_t}\in \mathbb{R}^{3N_D \times 3N_D}$ is calculated to  compute $\check{\boldsymbol{P}}_t$. For the second step the algorithm uses the measurements to iteratively update the values of the state and covariance in this step the jacobian  $\boldsymbol{H}=\tfrac{\partial h(\boldsymbol{x}_t)}{\partial \boldsymbol{x}_t}\in \mathbb{R}^{3N_D \times (6 N_I+3N_M+N_D)}$ is calculated to compute $\boldsymbol{K}_t$, $\hat{\boldsymbol{x}}_t$ and $\hat{\boldsymbol{P}}_t$. In this work, both Jacobians are calculated numerically by perturbations~\cite{NIANDONG2017388}.
To deal with the complex non-linear functions associated with the estimation problem, the algorithm refines the state predictions iteratively as shown in Figure~\ref{fig:Pipeline2} and Alg.~\ref{IEKF_OS_Algorithm}. The number of these iterations $\epsilon$ is tuned to optimize both the accuracy and the computational efficiency of the algorithm~\cite{Havlík_2015}. We assume that there is no cross-coupling between the predicted states. Hence, we assume an identity process noise distribution matrix $\boldsymbol{L}= \mathbf{I}_{3N_D\times3N_D} $. 

\begin{algorithm}
\caption{Tightly Coupled Multibody Kinematics and Kinetics estimation algorithm}
\label{IEKF_OS_Algorithm}
\begin{minipage}{\linewidth-18pt}
\begin{algorithmic}[1]
\REQUIRE Sensor measurements $\boldsymbol{y}_t$ shown in~\eqref{meas1}, an initial state estimate $\hat{x}_0$ with covariance $\hat{P}_0$, process and measurement noise covariances matrices $\boldsymbol{Q}$ and $\boldsymbol{R}$ demonstrated  in~\eqref{eq:stateSpaceModel} and~\eqref{meas1} respectively and the number of iterations $\epsilon$.

\ENSURE Estimates of the system states $\hat{x}_t$ and the respective covariance $\hat{P}_t$ for $t = 1, \hdots T$.\\

\STATE \textbf{for} $t = 1, \hdots, T$ \textbf{do}
\STATE \quad \textbf{Time update}

\quad \quad Compute the state prediction $\check{x}_t$   
            using~\eqref{eq:dynModel-discrete} and $\hat{x}_{t-1}$. Also compute the state\\ \quad\quad covariance $\boldsymbol{\check{P}}_t$ as
    \begin{align}
        \boldsymbol{\check{P}}_t &= \boldsymbol{G}_{t-1}\boldsymbol{\hat{P}}_{t-1} \boldsymbol{G}_{t-1}^\Transp + \boldsymbol{L} \boldsymbol{Q} \boldsymbol{L}^\Transp,
    \end{align}
   \quad\quad  with $\boldsymbol{G}_{t-1}=\left.\frac{\partial g(\dot{\boldsymbol{x}}_{t-1})}{\partial \boldsymbol{x}_{t-1}}\right|_{\boldsymbol{x}_{t-1}=\hat{\boldsymbol{x}}_{t-1}}$.
    

\STATE \quad \textbf{Measurement update}
\STATE \quad \quad \textbf{for} $k = 1, \hdots, \epsilon$ \textbf{do}
\STATE \quad \quad \quad Update the state with measurements $y_t$ as
            \begin{align}
            \bar{\boldsymbol{x}}_t^{k+1} &= \check{\boldsymbol{x}}_t + \boldsymbol{K}_t^k \bigg(y_t - {h}_t(\boldsymbol{\bar{x}}_t^k) - \boldsymbol{H}_t^k (\check{\boldsymbol{x}}_t - \bar{\boldsymbol{x}}_t^k)\bigg),
            \label{UpdateStateIEKF1}
        \end{align}
    \quad \quad \quad where $\boldsymbol{K}_t^k =\check{P}_t ( \boldsymbol{H}_t^k )^\Transp \left( \boldsymbol{H}_t^k \boldsymbol{\check{\boldsymbol{P}}}_t ( \boldsymbol{H}_t^k )^\Transp + \boldsymbol{R} \right)^{-1}$ and
    $\boldsymbol{H}_t^k =\left.\frac{\partial h_t(\boldsymbol{x}_t)}{\partial \boldsymbol{x}_t}\right|_{\boldsymbol{x}_{t}=\bar{\boldsymbol{x}}^k_t}$.
\STATE \quad \quad Compute the filtered state and covariance 
    \begin{align}
        \hat{\boldsymbol{x}}_t = \bar{\boldsymbol{x}}_t^{\epsilon+1}, \qquad 
        \hat{\boldsymbol{P}}_t = \Big ( \mathcal{\boldsymbol{I}} - \boldsymbol{K}_t^\epsilon \boldsymbol{H}_t^\epsilon \Big ) \check{\boldsymbol{P}}_t.
    \end{align}
\end{algorithmic}
\end{minipage}
\end{algorithm}

Both stationary and pre-known dynamic sensor measurements were recorded and then compared against ground truth measurements discussed further in~\ref{results1}. The error between them was used to compute the measurement covariance matrix $\boldsymbol{R}$. The number of iterations, 
$\epsilon$, was chosen experimentally when it was noticed that there was no significant variation between consecutive iterations beyond that value.

\subsection{Real world testing on physical systems}
\label{EXPsetups}
Two separate sets of experiments were performed on two different systems to validate our approach.
The experimental setups in this study were chosen to ensure accurate ground truth data to validate our proposed algorithm. Therefore, we omitted testing on humans, since precise ground truth for human motion cannot be reliably established. 
\begin{figure}
\centering
\includegraphics[width=0.32\linewidth, height=0.42\textheight]{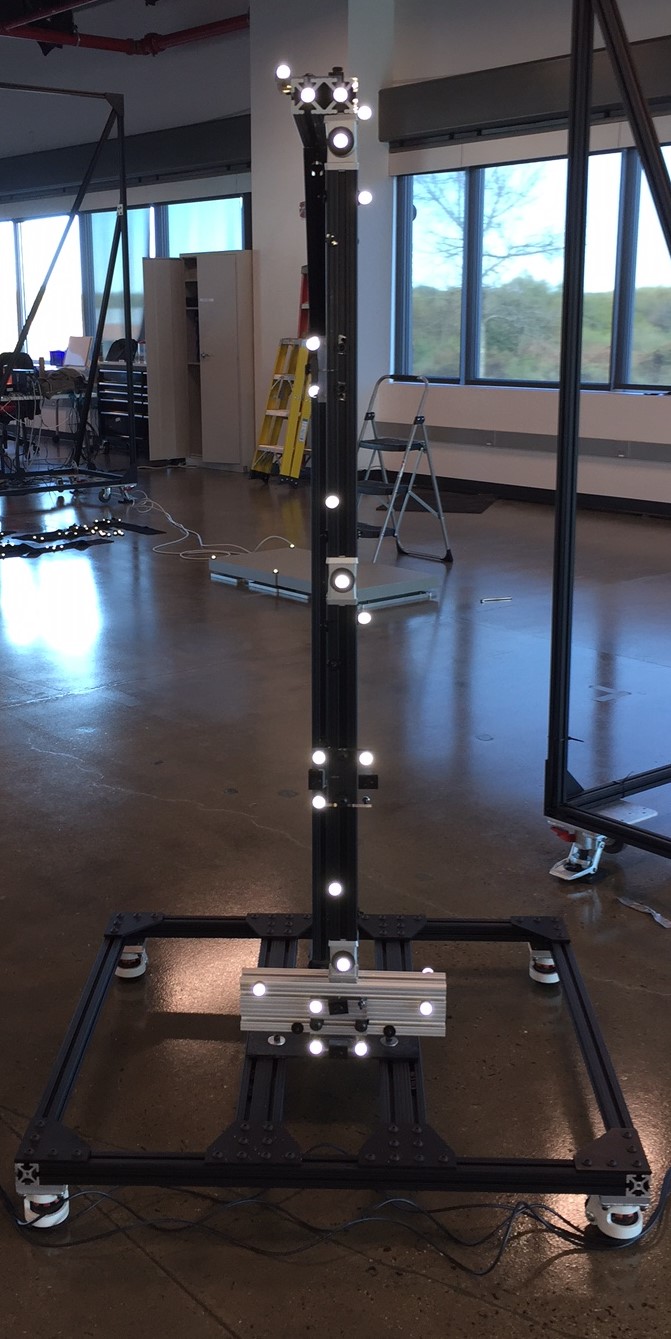}
\includegraphics[width=0.34\linewidth, height=0.42\textheight]{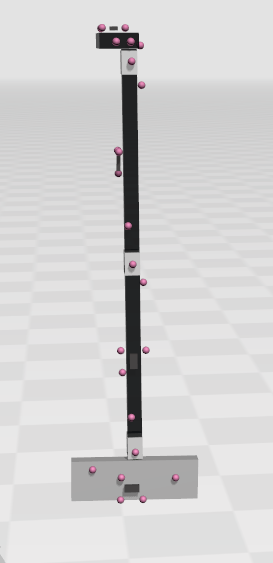}
\caption{An experimental setup with 3DOF is shown on the left, where optical markers and an IMU are attached to each body segment. On the right is an OpenSim model with corresponding IMU and marker placements. Sensor locations on the model were registered  using Optical MoCap measurements.}
\label{fig:Pend}
\end{figure}

The first experimental setup used was a 3DOF aluminum pendulum, shown in Figure~\ref{fig:Pend}. The pendulum consisted of four bodies and three pin joints.  Each body segment was equipped with an IMU and a set of active light-emitting markers. Experiments were conducted on this passive multibody system, in which the lowest body segment shown in silver in Figure~\ref{fig:Pend} was raised to a certain height and then released to swing under its own inertia. Our proposed approach is evaluated for estimating the pendulum motion with three different measurement models:
\begin{itemize} \item IMU measurements \item IMU and optical MoCap measurements \item IMU, optical MoCap, and virtual zero torque measurements \end{itemize} 
These measurement models demonstrate the ability of our algorithm to integrate different sensor measurements.
We compare the output of these measurement models with OpenSim marker-based and IMU-based (OpenSense) Inverse Kinematics (IK) and Inverse Dynamics (ID) across six trials, each with different initial generalized joint angles.

\begin{figure}
\centering
\includegraphics[width=0.23\linewidth, height=0.40\textheight]{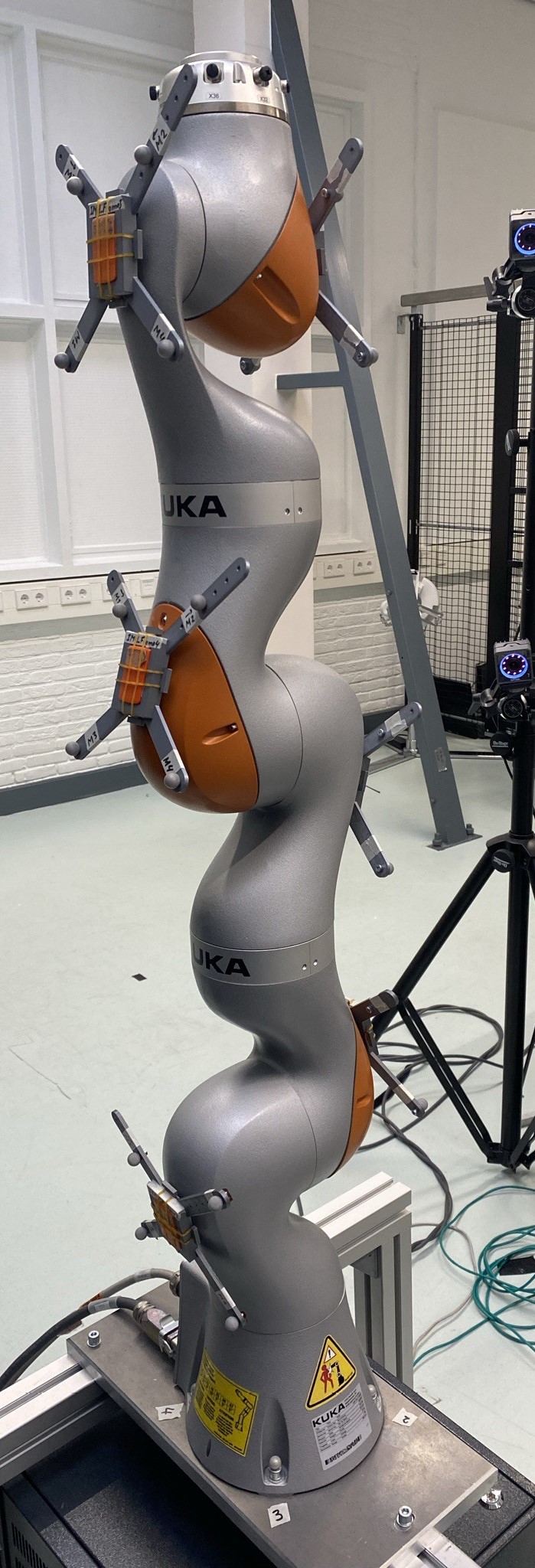}
  \hspace{0.0001cm}
\includegraphics[width=0.23\linewidth, height=0.40\textheight]
{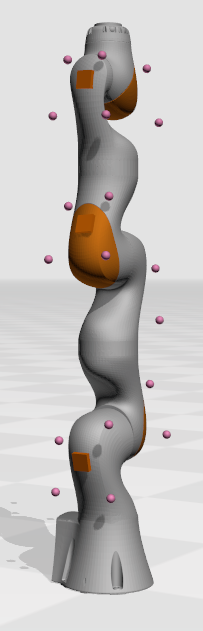}
\caption{On the left the actual KUKA LBR iiwa R800 that was used where a set of 4 markers and an IMU were placed on each body segment. On the right an OpenSim model that describes the system is shown.}
\label{KukaModel}
\end{figure}

The second experimental setup used was a KUKA LBR 7 iiwa R800 robot shown in Figure~\ref{KukaModel}. Apart from the end effector, the robot consists of 7 bodies and 6 revolute joints. Markers and IMUs were fitted on all the robot body segements except the base and the end effector. We use our approach to estimate the motion of the robot using two different measurement models. The IMU measurements only, and both the IMU measurements and the MoCap measurements. All IMUs were pre-calibrated before each trial in this experiment. 
 IK in this experiment was only conducted using the optical MoCap data. Due to the magnetic disturbances, performing IK using the IMU orientations was not feasible.
A total of 5 different trials were conducted, 1 using both the IMU and the MoCap measurements, and 4 using only the IMU measurements. 
\begin{figure}
    \centering
    \begin{subfigure}{\textwidth}
        \centering
        \includegraphics[width=.75\linewidth]{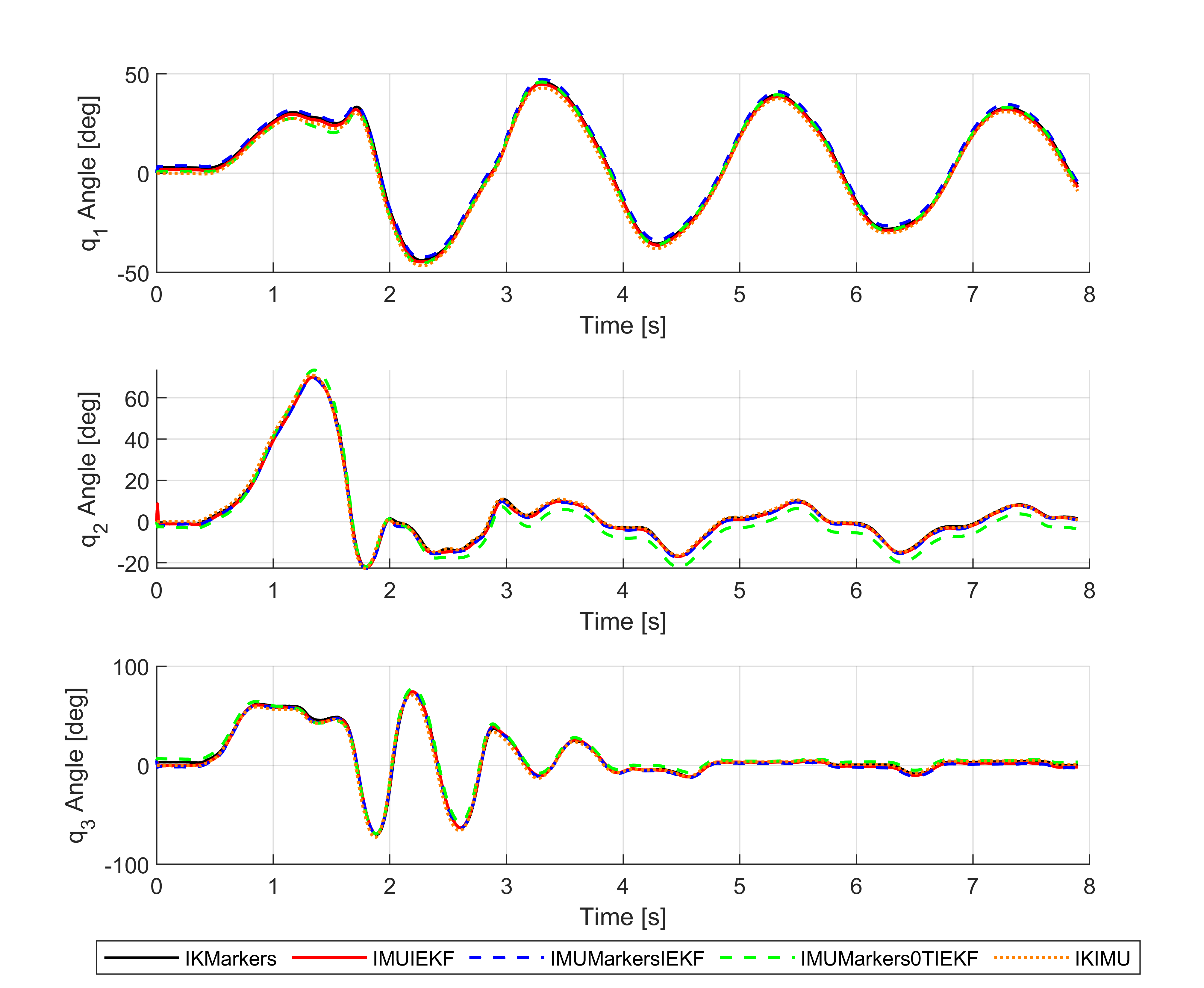}
    \end{subfigure}
       \begin{subfigure}{\textwidth}
        \centering
        \includegraphics[width=.75\linewidth]{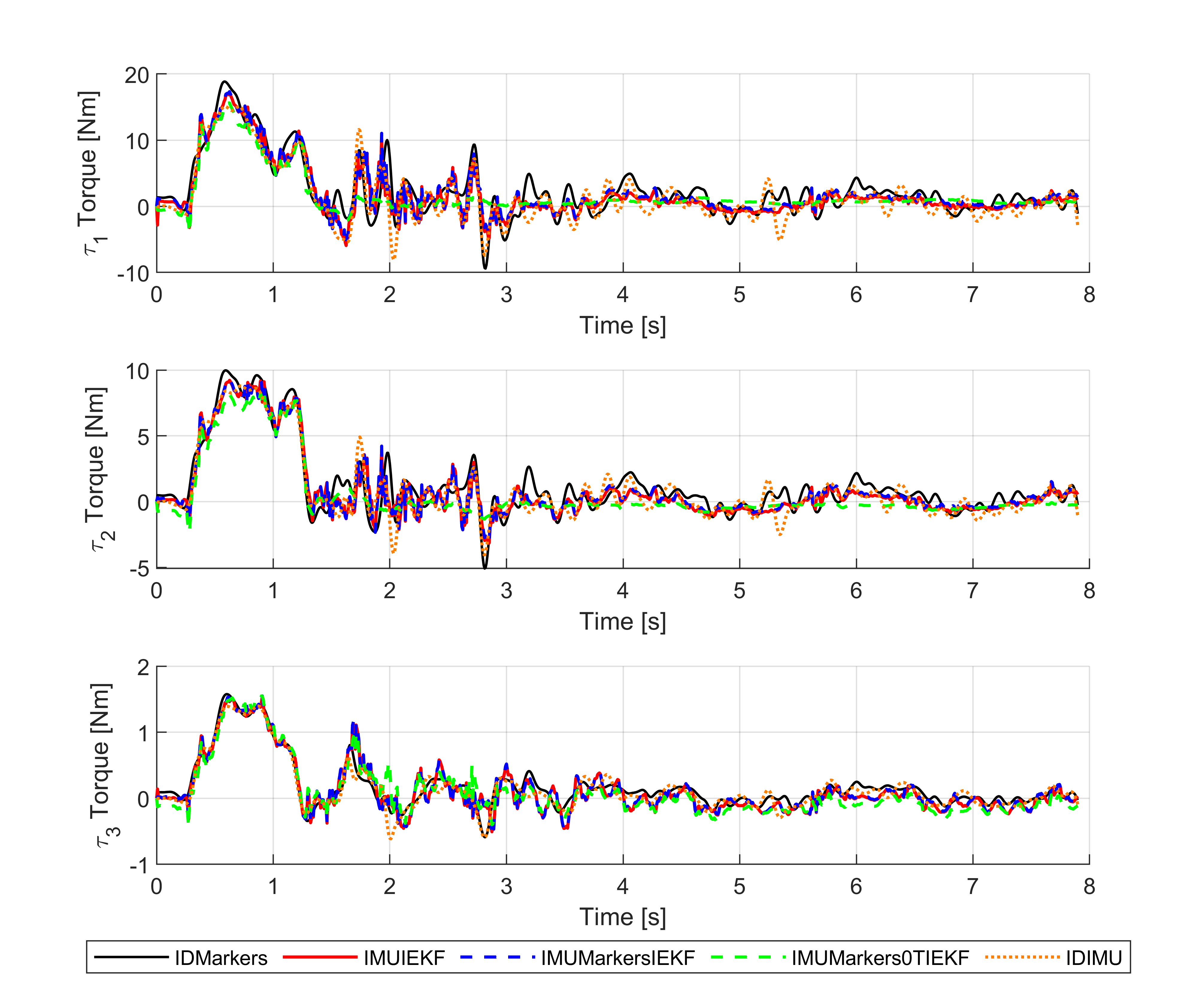}
    \end{subfigure}
    
    \caption{Shows trail 3 estimated generalized joint angles and torques of the 3DOF pendulum using the IMU data only, The IMU and Marker data and The IMU and Marker data with zero torque update all within the proposed tightly coupled IEKF approach. This is plotted against OpenSim's IK/ID.}
    \label{fig:exp4}
\end{figure}

\section{Results}
\label{results1}
In this section, we present the results and findings of the trials conducted on the two experimental setups discussed in~\ref{methodlgy}. For each experimental setup, we illustrate the output of the different measurement models by showing the estimated joint angles and torques plot against the groundtruth for one trial and the Root Mean Squared Difference (RMSD) values for all the trails.

Figure~\ref{fig:exp4} presents the estimated joint angles and torques for trial 3 conducted on the pendulum experimental setup using different measurement models. At time 0 s, the pendulum was in its initial position shown in Figure~\ref{fig:Pend}. the pendulum was raised and then left to swing freely under its own inertia from time 1.2 s to 7.8 s. The plots in the figure indicates that the joint angle estimates from the different measurement models closely follow the same trend. However, a steady offset of approximately \(2^\circ\) is observed in the IMU marker-zero torque update $q_2$ estimates after 2.4 s compared with the other measurement models. The joint torque estimates from the different measurement models also appear to follow the same trend, although each model exhibits a slightly different oscillation pattern. Notably, both the marker and IMU-based Inverse Dynamics (ID) estimates show higher peaks around zero after time 3 s in both \(\tau_1\) and \(\tau_2\).

Table~\ref{tab:exp_q_tau_all} presents the generalized joint angles and torques RMSD values of the different measurement models compared to the marker based IK and ID for the six pendulum trials. The table also shows RMSD values of the different measurement models compared with the IMU marker IEKF estimates.  In trial 1, the estimate of IMU-based IK (OpenSense-IK) for \( q_3 \) had an RMSD of \( 6.56^\circ \) compared to \( 3.49^\circ \) achieved by IMU-IEKF, indicating a 46.8\% marker-IK RMSD reduction. Similarly, in trials 2, 3, and 4, the IMU-IEKF demonstrated RMSD reductions of 31.7\%, 48.6\%, and 18.6\%, respectively, over the OpenSense-IK estimate for \( q_3 \). While in trail 6 OpenSense-IK estimate for \( q_3 \) had an RMSD of \( 3.23^\circ \) compared to \( 3.72^\circ \) achieved by IMU-IEKF showing a reduction of 13.17\%. The RMSD values of both the IMU-marker IEKF and IMU-IEKF, calculated with respect to the marker-IK estimates, are closely matched, with a maximum difference of \( 0.64^\circ \). This difference reflects a 17.2\% reduction in RMSD for the IMU-marker IEKF \( q_3 \) estimate in trial 6.

As for the torque Marker-ID RMSD, the maximum difference between OpenSense-ID and IMU-IEKF was 1.62 Nm in $\tau_1$ during trail 6. The IMU-IEKF and IMU-Marker IEKF RMSD values were close with a maximum difference of 0.10 Nm in $\tau_1$ during trail 6.

\begin{table}
\centering
\renewcommand{\arraystretch}{1.1}
\setlength{\tabcolsep}{5pt}
\begin{tabular}{llccccc}
\toprule
\textbf{T} & \textbf{Metric} & \textbf{IMU} & \textbf{IMU Markers} & \textbf{IMU Markers 0T} & \textbf{OS} & \textbf{Markers} \\
\midrule
\multirow{6}{*}{1} 
 & $q_1$    & 1.27\,|\,0.10 & 1.28\,|\,-- & 1.29\,|\,0.10 & 0.92\,|\,2.13 & --\,|\,1.28 \\
 & $q_2$    & 0.91\,|\,0.12 & 0.93\,|\,-- & 0.99\,|\,0.09 & 0.92\,|\,1.28 & --\,|\,0.93 \\
 & $q_3$    & 3.49\,|\,0.12 & 3.54\,|\,-- & 4.23\,|\,0.93 & 6.56\,|\,3.32 & --\,|\,3.54 \\
 & $\tau_1$ & 1.05\,|\,0.03 & 1.07\,|\,-- & 0.92\,|\,0.47 & 0.78\,|\,1.56 & --\,|\,1.07 \\
 & $\tau_2$ & 0.39\,|\,0.01 & 0.40\,|\,-- & 0.34\,|\,0.28 & 0.39\,|\,0.66 & --\,|\,0.40 \\
 & $\tau_3$ & 0.07\,|\,0.00 & 0.07\,|\,-- & 0.11\,|\,0.07 & 0.15\,|\,0.16 & --\,|\,0.07 \\
\midrule
\multirow{6}{*}{2} 
 & $q_1$    & 1.10\,|\,0.11 & 1.11\,|\,-- & 1.11\,|\,0.10 & 1.54\,|\,2.26 & --\,|\,1.11 \\
 & $q_2$    & 0.94\,|\,0.17 & 0.97\,|\,-- & 1.02\,|\,0.11 & 2.21\,|\,2.60 & --\,|\,0.97 \\
 & $q_3$    & 3.75\,|\,0.10 & 3.72\,|\,-- & 4.16\,|\,0.75 & 5.50\,|\,4.05 & --\,|\,3.72 \\
 & $\tau_1$ & 1.70\,|\,0.05 & 1.70\,|\,-- & 1.33\,|\,0.82 & 1.70\,|\,1.91 & --\,|\,1.70 \\
 & $\tau_2$ & 1.01\,|\,0.01 & 1.00\,|\,-- & 0.79\,|\,0.48 & 0.75\,|\,1.09 & --\,|\,1.00 \\
 & $\tau_3$ & 0.18\,|\,0.00 & 0.18\,|\,-- & 0.21\,|\,0.12 & 0.12\,|\,0.26 & --\,|\,0.18 \\
\midrule
\multirow{6}{*}{3} 
 & $q_1$    & 1.34\,|\,0.10 & 1.35\,|\,-- & 1.38\,|\,0.10 & 1.00\,|\,1.67 & --\,|\,1.35 \\
 & $q_2$    & 0.76\,|\,0.16 & 0.79\,|\,-- & 0.87\,|\,0.13 & 1.10\,|\,1.27 & --\,|\,0.79 \\
 & $q_3$    & 1.64\,|\,0.09 & 1.66\,|\,-- & 1.86\,|\,0.40 & 3.18\,|\,2.81 & --\,|\,1.66 \\
 & $\tau_1$ & 1.23\,|\,0.05 & 1.24\,|\,-- & 1.25\,|\,0.66 & 1.45\,|\,1.52 & --\,|\,1.24 \\
 & $\tau_2$ & 0.63\,|\,0.01 & 0.63\,|\,-- & 0.62\,|\,0.39 & 0.64\,|\,0.71 & --\,|\,0.63 \\
 & $\tau_3$ & 0.14\,|\,0.00 & 0.14\,|\,-- & 0.18\,|\,0.09 & 0.06\,|\,0.16 & --\,|\,0.14 \\
\midrule
\multirow{6}{*}{4} 
 & $q_1$    & 1.28\,|\,0.13 & 1.27\,|\,-- & 1.29\,|\,0.11 & 1.38\,|\,2.57 & --\,|\,1.27 \\
 & $q_2$    & 1.19\,|\,0.38 & 1.11\,|\,-- & 1.22\,|\,0.13 & 0.65\,|\,1.07 & --\,|\,1.11 \\
 & $q_3$    & 2.75\,|\,0.28 & 2.69\,|\,-- & 3.08\,|\,0.56 & 3.38\,|\,3.91 & --\,|\,2.69 \\
 & $\tau_1$ & 1.53\,|\,0.18 & 1.52\,|\,-- & 1.42\,|\,0.71 & 1.34\,|\,2.01 & --\,|\,1.52 \\
 & $\tau_2$ & 0.76\,|\,0.05 & 0.76\,|\,-- & 0.69\,|\,0.43 & 0.60\,|\,0.89 & --\,|\,0.76 \\
 & $\tau_3$ & 0.14\,|\,0.00 & 0.14\,|\,-- & 0.18\,|\,0.10 & 0.08\,|\,0.18 & --\,|\,0.14 \\
\midrule
\multirow{6}{*}{5} 
 & $q_1$    & 1.05\,|\,0.12 & 1.04\,|\,-- & 1.05\,|\,0.11 & 0.86\,|\,1.51 & --\,|\,1.04 \\
 & $q_2$    & 1.24\,|\,0.58 & 1.18\,|\,-- & 1.18\,|\,0.18 & 2.09\,|\,2.29 & --\,|\,1.18 \\
 & $q_3$    & 2.67\,|\,0.68 & 2.65\,|\,-- & 2.94\,|\,0.56 & 2.37\,|\,2.60 & --\,|\,2.65 \\
 & $\tau_1$ & 1.85\,|\,0.07 & 1.85\,|\,-- & 1.74\,|\,0.69 & 1.18\,|\,2.06 & --\,|\,1.85 \\
 & $\tau_2$ & 1.02\,|\,0.07 & 1.02\,|\,-- & 0.95\,|\,0.41 & 0.60\,|\,1.10 & --\,|\,1.02 \\
 & $\tau_3$ & 0.26\,|\,0.02 & 0.26\,|\,-- & 0.26\,|\,0.09 & 0.05\,|\,0.26 & --\,|\,0.26 \\
\midrule
\multirow{6}{*}{6} 
 & $q_1$    & 1.45\,|\,0.80 & 1.27\,|\,-- & 1.29\,|\,0.19 & 1.09\,|\,1.48 & --\,|\,1.27 \\
 & $q_2$    & 1.75\,|\,0.43 & 1.64\,|\,-- & 1.67\,|\,0.24 & 1.43\,|\,1.51 & --\,|\,1.64 \\
 & $q_3$    & 3.72\,|\,1.64 & 3.08\,|\,-- & 3.04\,|\,1.11 & 3.23\,|\,2.80 & --\,|\,3.08 \\
 & $\tau_1$ & 3.02\,|\,0.52 & 2.92\,|\,-- & 2.51\,|\,0.84 & 1.40\,|\,2.85 & --\,|\,2.92 \\
 & $\tau_2$ & 1.72\,|\,0.22 & 1.68\,|\,-- & 1.47\,|\,0.50 & 0.64\,|\,1.59 & --\,|\,1.68 \\
 & $\tau_3$ & 0.48\,|\,0.03 & 0.48\,|\,-- & 0.47\,|\,0.10 & 0.09\,|\,0.49 & --\,|\,0.48 \\
\bottomrule
\end{tabular}
\caption{Joint angle RMSD (\textdegree) and joint torque RMSD (Nm) with respect to marker-based inverse kinematics (IK) and inverse dynamics (ID), as well as IMU-Marker IEKF, across six trials. Columns: \textbf{IMU} = IMU-only IEKF, \textbf{IMU Markers} = IMU markers IEKF, \textbf{IMU Markers 0T} = IMU markers with zero torque updates IEKF, \textbf{OS} = OpenSense, \textbf{Markers} = marker-based IK or ID. Values are formatted as \textit{marker-based RMSD\,|\,IMU-Marker IEKF RMSD}.}
\label{tab:exp_q_tau_all}
\end{table}
\begin{figure}
    \centering
    \begin{subfigure}{\textwidth}
        \centering
        \includegraphics[width=.75\linewidth]{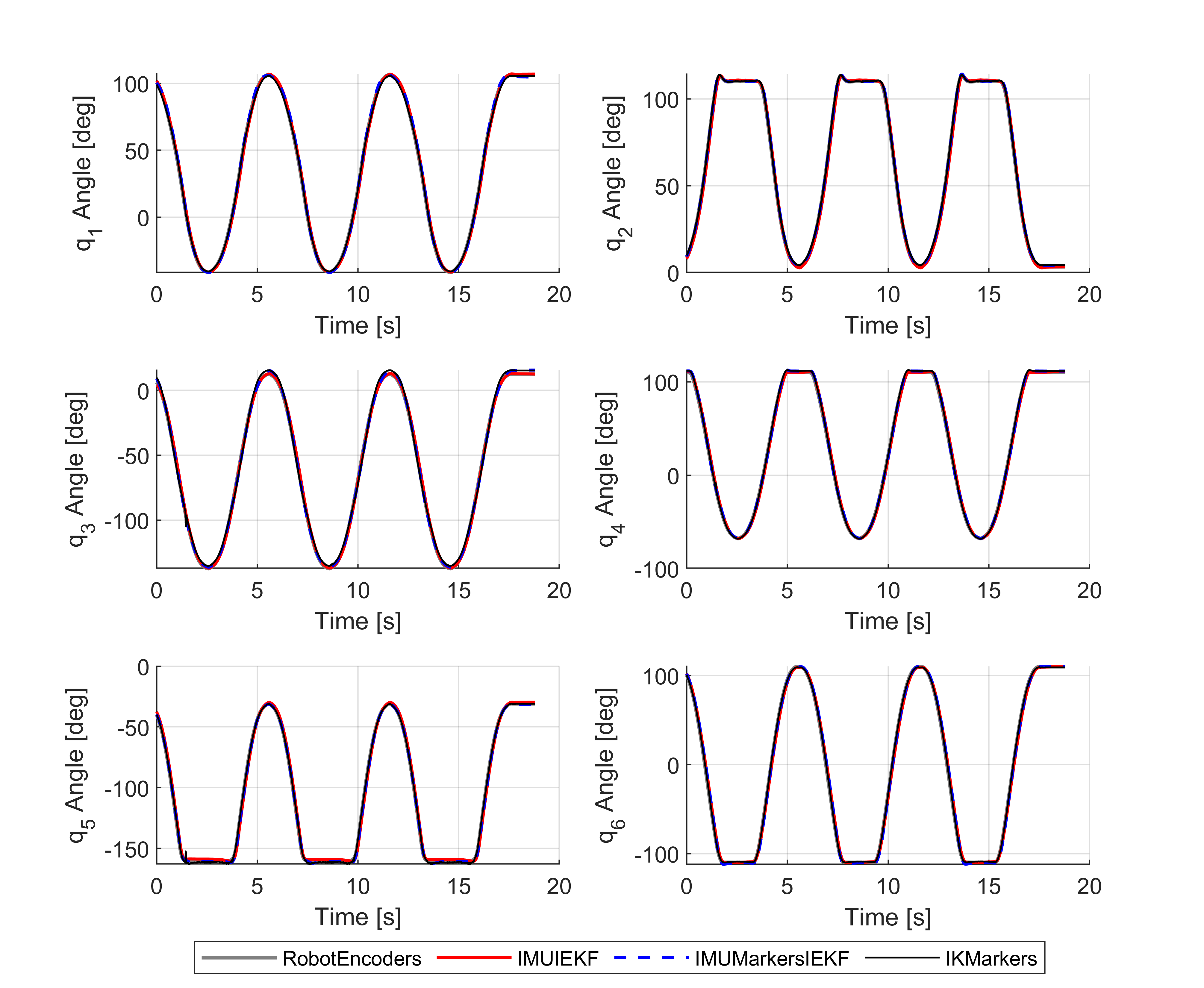}
    \end{subfigure}
       \begin{subfigure}{\textwidth}
        \centering
        \includegraphics[width=.75\linewidth]{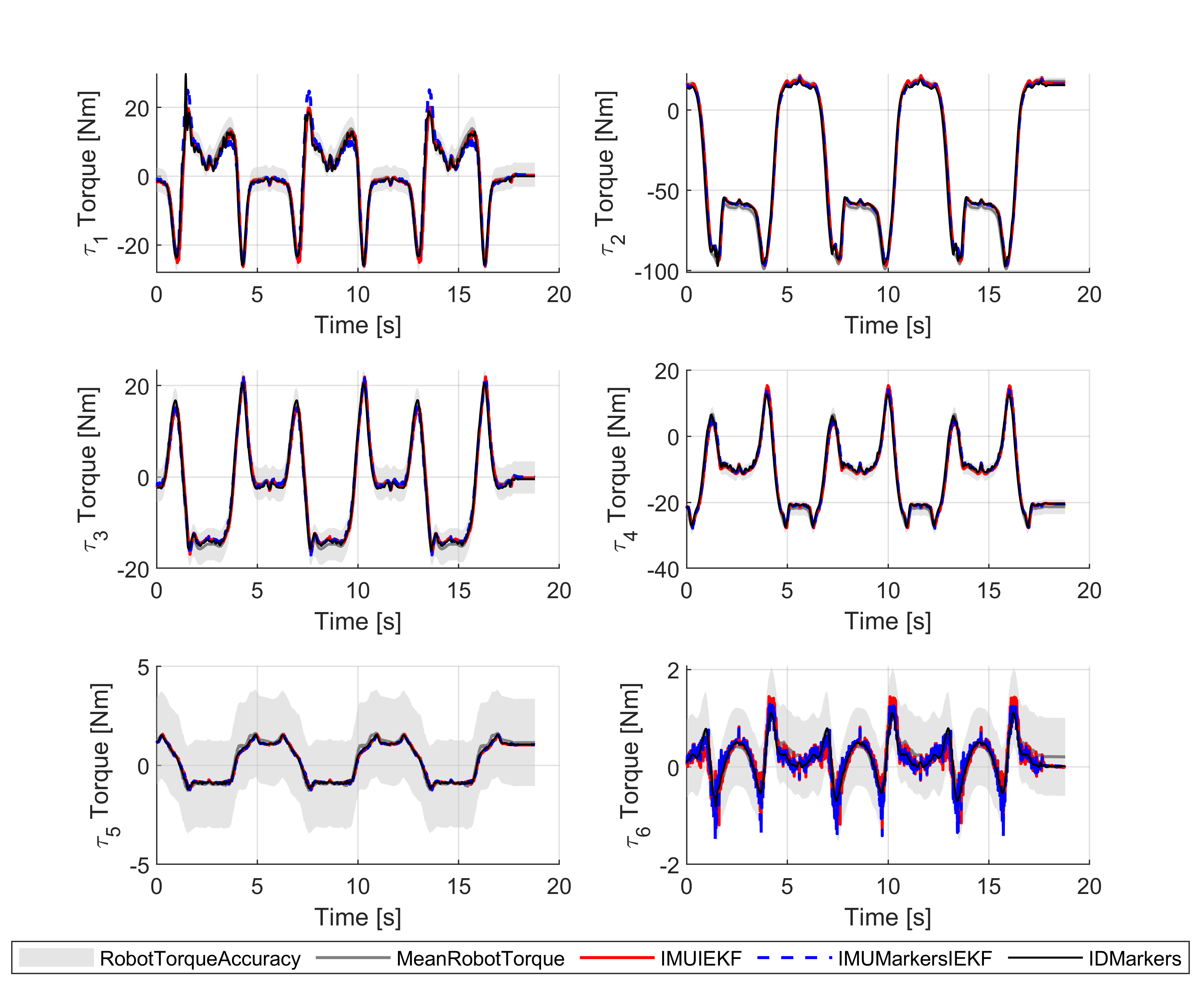}
    \end{subfigure}
     \caption{Shows trail 1 estimated joint angles and torques using our proposed IMU-IEKF measurment model, the IMU-Marker IEKF. This is plotted against OpenSim's IK/ID and the robot's own encoders and torque sensors.}
    \label{fig:exp1}
    \end{figure}

\begin{figure}[ht]
    \centering
    \includegraphics[width=0.4\linewidth]{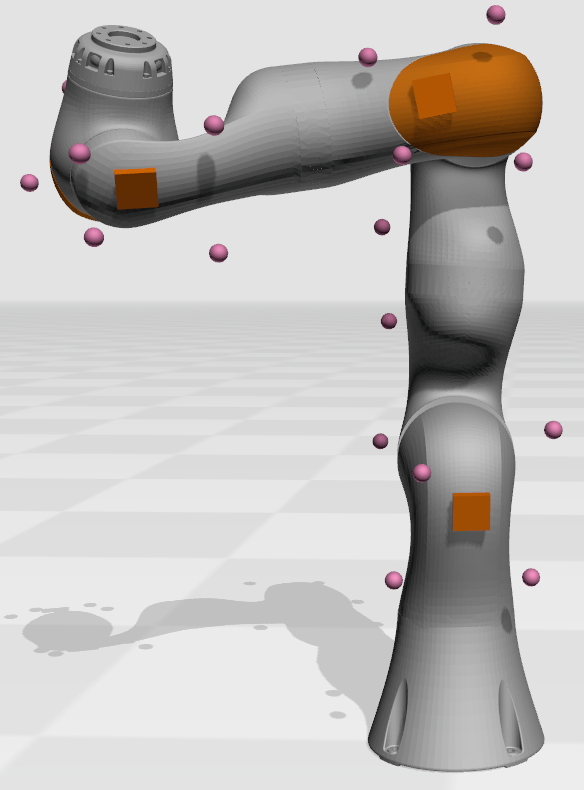}
    \caption{Initial pose of the Kuka in Trial 1.}
    \label{fig:trail11}
\end{figure}

\begin{figure}
    \centering
    \includegraphics[width=0.85\linewidth, height=0.45\textheight]{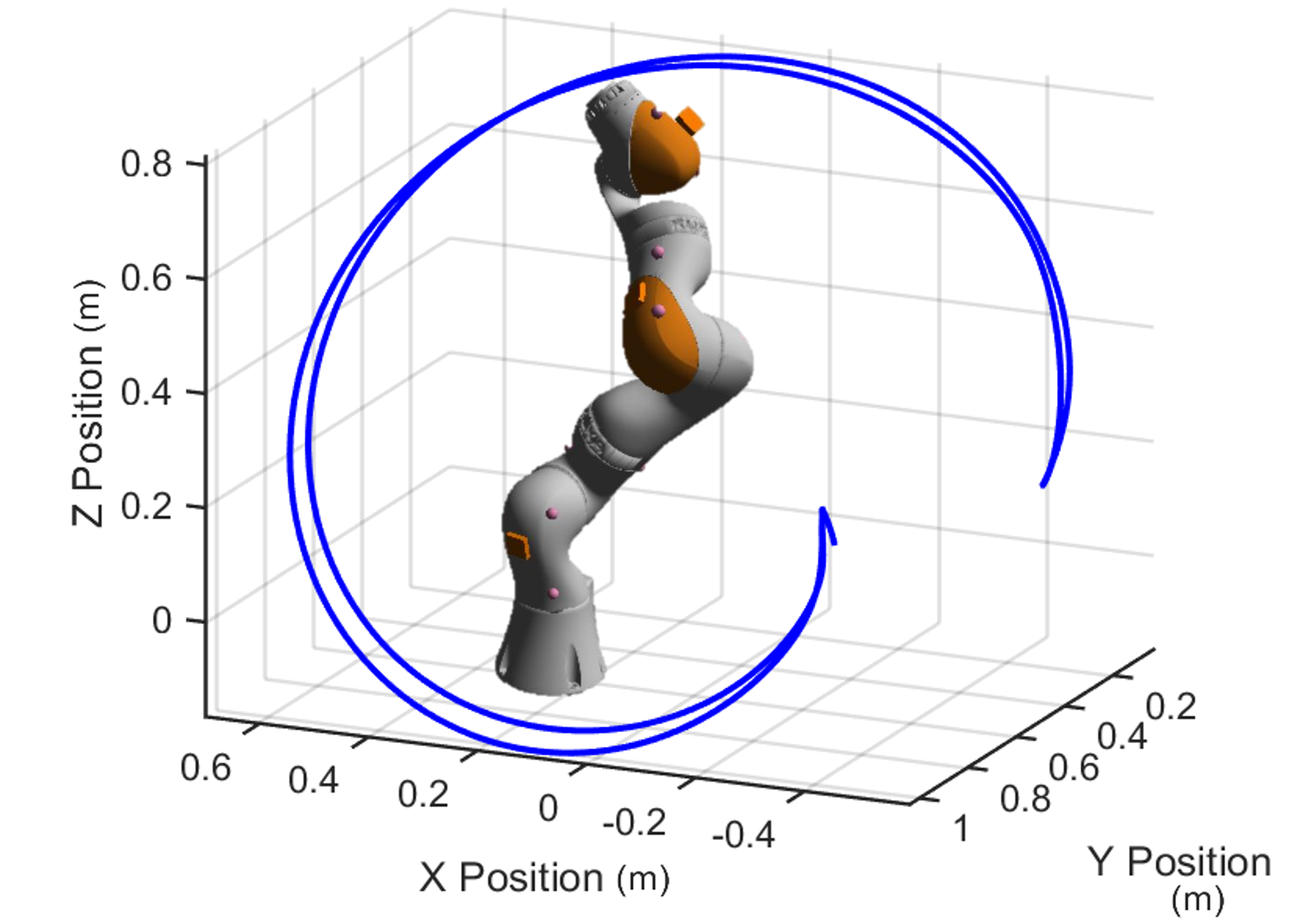}
    \caption{3D visualization of the end effector motion path for the Kuka Robot trail 1.}
    \label{fig:trajectory}
\end{figure}

Figure~\ref{fig:exp1} presents the estimated generalized joint angles and torques for the Kuka robot trial 1. The initial pose of the robot during this trial is shown in Figure~\ref{fig:trail11}. The total trial time was about 18 s where all joints were  simultaneously actuated except for the end effector. The end effector motion path of this trial is shown in figure~\ref{fig:trajectory}. The estimated joint angles of the different measurement models shown in the figure appear to have small differences. However, the RMSD values calculated with respect to the Kuka encoders in Table~\ref{tab:combined_rmsd} show a maximum RMSD of \( 3.24^\circ \) for the IMU-IEKF $q_6$ estimate, \( 2.84^\circ \) for the IMU-Marker IEKF $q_6$ estimate and \( 1.16^\circ \) for the marker-IK $q_3$ estimate. The joint torque estimates of the different models also show small differences. However, The Marker ID $\tau_1$  estimate shows a difference of about 4.2 Nm compared to the other models at time 1.2 s. 
The RMSD values calculated with respect to the Kuka torque sensors show a maximum of 4.27 Nm for IMU-IEKF, $3.71$ Nm for IMU-Marker IEKF and 2.97 Nm in the Marker ID $\tau_2$ estimates.

Table~\ref{tab:combined_rmsd} shows both the estimated joint angles and torques for trials 2 to 5 conducted on the Kuka robot. These RMSD values are calculated with respect to the IMU-IEKF estimates and the robot encoders and torque sensors measurements. The total time of these trials ranged between 60 s for trial 2 and 30 s for trial 5. The joint angles RMSD values for these experiments ranged between \( 3.01^\circ \) in trial 2 $q_3$ estimate and \( 0.39^\circ \) in trial 3 $q_2$ estimate. As for the joint torques the RMSD values ranged between 2.33 Nm in trial 2 $\tau_2$ estimate and 0.07 Nm in trial 4 and 5 $\tau_5$ estimate.

\begin{table}
\centering
\renewcommand{\arraystretch}{1.2}
\setlength{\tabcolsep}{5pt}
\begin{tabular}{lccc}
\toprule
\textbf{Joint} & \multicolumn{3}{c}{\textbf{Angle RMSD (\textdegree) / Torque RMSD (Nm)}} \\
\cmidrule(lr){2-4}
& \textbf{IMU-IEKF} & \textbf{IMU-M-IEKF} & \textbf{Marker-IK/ID} \\
\midrule
$q_1$|$\tau_1$ & 1.76 | 1.99 & 1.37 | 2.23 & 0.33 | 1.67 \\
$q_2$|$\tau_2$ & 1.54 | 4.27 & 0.84 | 3.71 & 0.66 | 2.97 \\
$q_3$|$\tau_3$ & 1.44 | 1.41 & 1.45 | 1.13 & 1.16 | 1.01 \\
$q_4$|$\tau_4$ & 2.31 | 1.64 & 1.45 | 1.51 & 0.75 | 1.25 \\
$q_5$|$\tau_5$ & 2.20 | 0.35 & 1.22 | 0.34 & 0.48 | 0.33 \\
$q_6$|$\tau_6$ & 3.24 | 0.37 & 2.84 | 0.38 & 0.36 | 0.33 \\
\bottomrule
\end{tabular}
\caption{Kuka experiment 1 joint angles and torques RMSD between different measurement models and the robot encoders and torque sensors.}
\label{tab:combined_rmsd}
\end{table}

\begin{table}
\centering
\renewcommand{\arraystretch}{1.2}
\setlength{\tabcolsep}{6pt} 
\begin{tabular}{lcccc}
\toprule
\textbf{Joint} & \multicolumn{4}{c}{\textbf{Angle RMSD (\textdegree) / Torque RMSD (Nm)}}\\
\cmidrule(lr){2-5} 
& \textbf{Trial 2} & \textbf{Trial 3} & \textbf{Trial 4} & \textbf{Trial 5} \\
\midrule
$q_1$|$\tau_1$ & 2.72 | 0.66  & 0.99 | 0.64  & 1.23 | 0.46  & 1.58 | 0.46  \\
$q_2$|$\tau_2$ & 1.86 | 2.33  & 0.39 | 1.38  & 0.87 | 1.65  & 0.43 | 1.81  \\
$q_3$|$\tau_3$ & 3.01 | 0.82  & 2.00 | 0.46  & 2.12 | 0.33  & 2.24 | 0.34  \\
$q_4$|$\tau_4$ & 2.06 | 0.45  & 0.77 | 0.32  & 1.24 | 0.36  & 0.95 | 0.38  \\
$q_5$|$\tau_5$ & 1.52 | 0.08  & 1.57 | 0.08  & 0.86 | 0.07  & 1.00 | 0.07  \\
$q_6$|$\tau_6$ & 1.73 | 0.23  & 1.42 | 0.23  & 1.77 | 0.22  & 1.26 | 0.23  \\
\bottomrule
\end{tabular}
\caption{Kuka trials 2 to 5 joint angles  and torques RMSD between IMU based IEKF and the robot encoders and torque sensors.}
\label{tab:experiment_rmsd}
\end{table}

\section{Discussion}

The results of the trials conducted on the pendulum experimental setup indicate that our algorithm generally provided more accurate joint angle estimates than the OpenSense approach. The maximum joint angle RMSD for our algorithm did not exceed \(3.75^\circ\), while the OpenSense approach reached a maximum RMSD of \(6.57^\circ\), exceeding the \(5^\circ\)  RMSD benchmark suggested for human upper extremity analysis~\cite{Morrow2017,Fang2023}. However, in some trials, the OpenSense estimates showed a reduction in RMSD errors over the IMU-IEKF. Although this reduction did not exceed \(0.48^\circ\), it may be attributed to a slight misalignment in the sensors translation registration. Since, OpenSense primarily depends on an accurate orientation alignment only, which was the same for both approaches. The results also indicated that the zero torque virtual measurements slightly impacted the accuracy of the generalized joint angles. This effect is likely due to the presence of dynamic friction in the joint bearings, which
introduces non-zero torques.
Across all 6 trials, torque estimates from the various measurement models showed strong agreement. In contrast to methods that rely on ID, which typically require low-pass filtering to mitigate noise. Our approach provides estimates without being affected by such processing steps. As shown in Figure~\ref{fig:exp4}, the measurement models, particularly the zero-torque update IEKF, closely converge toward zero torque from around 3 s onward. This behavior aligns well with the expected physical dynamics of the pendulum, where minimal torque is anticipated during this phase of the motion.

The Kuka results show the ability of our approach to provide accurate motion estimates in environments of high magnetic distortions where this was not possible with the OpenSense approach. The 5 trials motion paths were chosen to utilize the high maneuverability of the Kuka robot, demonstrating the ability of our proposed approach to track the corresponding kinematics and kinetics. The RMSD values presented in Table~\ref{tab:combined_rmsd} and Table~\ref{tab:experiment_rmsd} show joint angles RMSD that are consistent with the pendulum trials with values not exceeding $3.24^\circ$. As for the torque estimates the results illustrates the ability of the IMU-IEKF to closely follow the trajectory providing torque estimates that are aligned with robot's torque sensors given that the Kuka's torque sensors have an accuracy of $\pm2\%$ of the maximum torque which ranges between 176 Nm and 40 Nm for joints 1 to joint 6~\cite{kuka_iiwa_torque,kuka_iiwa_accuracy}. This torque accuracy is shown in Figure~\ref{fig:exp1} where it can be noticed that nearly all the torque estimates were between the accuracy bounds of the torque sensors shown in gray. Figure~\ref{fig:exp1} demonstrates that our proposed pipeline is able to track the abrupt changes in the torque profile accurately. In $\tau_1$ estimate at time 1.2 s the already filtered inverse dynamics had an overshoot of 4.2 Nm. However, both the IMU-IEKF and the IMU-Marker IEKF were able to track the robot's torque sensor measurements. The maximum torque RMSD was 4.27 Nm
which constitutes approximately $2.5\%$ of the maximum joint torque which is comparable to the $\pm2\%$ accuracy of the Kuka's torque sensor. 
The inclusion of markers in the measurement model improved accuracy, by a 15\%-40\% reduction in the joint angles RMSD values when using the IMU-Marker IEKF compared with the IMU-IEKF. these results show the effectiveness of combining different measurement models to enhance accuracy. Although the marker-based IK and ID estimates show lower RMSD values than the IMU-IEKF, it's important to note the accuracy of the torque sensor. Additionally, IMUs are more cost-effective and can be used in a wider range of environments and settings.

Certain assumptions were made in this work, highlighting potential limitations. First, the locations of the IMUs were assumed to be rigid (i.e., no soft tissue artifacts) and accurately known a priori, which may not always hold in practical applications. Second, the process noise $\boldsymbol{Q}$ should ideally be estimated based on the model and the current system state, but this remains challenging. The highly nonlinear nature of the system model complicates the computation of the measurement model Jacobian $H_t$. Although the IEKF mitigates these challenges by iterating the linearization process, this results in a computationally expensive algorithm.

Future improvements could address these limitations by incorporating an automatic  sensor registration algorithm, which would solve for the sensor orientation and translation with respect to the body segment using the model kinematic properties in addition to the sensors measurements. The algorithm can also include an adaptive process and measurement covariance estimator to improve the estimation accuracy specially if applied on different multibody systems in different environments. Additionally, estimating the gyroscope bias within the system states could enhance accuracy. The use of a differentiable model could also improve Jacobian calculation, potentially balancing both accuracy and computational efficiency.
\subsection{conclusions and future work}
In this work, we presented a tightly coupled IMU-based motion capture (MoCap) algorithm that integrates various measurement models to estimate both the kinematics and kinetics of a multibody system using an IEKF and the system’s dynamic model. The proposed approach was validated using two different multibody systems: a non-actuated pendulum and an actuated robotic arm. Our results demonstrated that the algorithm can accurately estimate both kinematics and kinetics solely from IMU measurements. This capability enables motion capture beyond constrained laboratory environments, facilitating accessible human kinematic and kinetic estimation.

Integrating system kinematics and dynamic constraints into the estimation process was shown not only to enhance accuracy but also to mitigate the effects of omitting magnetometer measurements, thereby ensuring reliable estimates even in environments with strong magnetic disturbances. Additionally, the algorithm's ability to integrate multiple sensors further improved estimation accuracy.

\bibliography{sample}

\begin{thebibliography}{10}
\urlstyle{rm}
\expandafter\ifx\csname url\endcsname\relax
  \def\url#1{\texttt{#1}}\fi
\expandafter\ifx\csname urlprefix\endcsname\relax\def\urlprefix{URL }\fi
\expandafter\ifx\csname doiprefix\endcsname\relax\def\doiprefix{DOI: }\fi
\providecommand{\bibinfo}[2]{#2}
\providecommand{\eprint}[2][]{\url{#2}}

\bibitem{menache2010understanding}
\bibinfo{author}{Menache, A.}
\newblock \emph{\bibinfo{title}{Understanding Motion Capture for Computer Animation}} (\bibinfo{publisher}{Morgan Kaufmann}, \bibinfo{year}{2010}).

\bibitem{MOESLUND200690}
\bibinfo{author}{Moeslund, T.~B.}, \bibinfo{author}{Hilton, A.} \& \bibinfo{author}{Krüger, V.}
\newblock \bibinfo{journal}{\bibinfo{title}{A survey of advances in vision-based human motion capture and analysis}}.
\newblock {\emph{\JournalTitle{Computer Vision and Image Understanding}}}  (\bibinfo{year}{2006}).

\bibitem{Suo24}
\bibinfo{author}{Suo, X.}, \bibinfo{author}{Tang, W.} \& \bibinfo{author}{Li, Z.}
\newblock \bibinfo{journal}{\bibinfo{title}{Motion capture technology in sports scenarios: A survey}}.
\newblock {\emph{\JournalTitle{Sensors}}}  (\bibinfo{year}{2024}).

\bibitem{LIAO2020}
\bibinfo{author}{Liao, Y.}, \bibinfo{author}{Vakanski, A.}, \bibinfo{author}{Xian, M.}, \bibinfo{author}{Paul, D.} \& \bibinfo{author}{Baker, R.}
\newblock \bibinfo{journal}{\bibinfo{title}{A review of computational approaches for evaluation of rehabilitation exercises}}.
\newblock {\emph{\JournalTitle{Computers in Biology and Medicine}}}  (\bibinfo{year}{2020}).

\bibitem{Ben2009}
\bibinfo{author}{van Basten, B. J.~H.}, \bibinfo{author}{Jansen, S. E.~M.} \& \bibinfo{author}{Karamouzas, I.}
\newblock \bibinfo{title}{Exploiting motion capture to enhance avoidance behaviour in games}.
\newblock In \emph{\bibinfo{booktitle}{Motion in Games}} (\bibinfo{publisher}{Springer Berlin Heidelberg}, \bibinfo{year}{2009}).

\bibitem{Salisu23}
\bibinfo{author}{Salisu, S.} \emph{et~al.}
\newblock \bibinfo{journal}{\bibinfo{title}{Motion capture technologies for ergonomics: A systematic literature review}}.
\newblock {\emph{\JournalTitle{Diagnostics}}}  (\bibinfo{year}{2023}).

\bibitem{Das2023}
\bibinfo{author}{Das, K.}, \bibinfo{author}{de~Paula~Oliveira, T.} \& \bibinfo{author}{Newell, J.}
\newblock \bibinfo{journal}{\bibinfo{title}{Comparison of {Markerless} and {Marker-Based} {Motion} {Capture} {Technologies} through {Simultaneous} {Data} {Collection} during {Gait}: {Proof} of {Concept}}}.
\newblock {\emph{\JournalTitle{Scientific Reports}}}  (\bibinfo{year}{2023}).

\bibitem{Liu2020}
\bibinfo{author}{Liu, S.}, \bibinfo{author}{Zhang, J.}, \bibinfo{author}{Zhang, Y.} \& \bibinfo{author}{Zhu, R.}
\newblock \bibinfo{journal}{\bibinfo{title}{A wearable motion capture device able to detect dynamic motion of human limbs}}.
\newblock {\emph{\JournalTitle{Nature Communications}}}  (\bibinfo{year}{2020}).

\bibitem{Fang23}
\bibinfo{author}{Fang, Z.}, \bibinfo{author}{Woodford, S.}, \bibinfo{author}{Senanayake, D.} \& \bibinfo{author}{Ackland, D.}
\newblock \bibinfo{journal}{\bibinfo{title}{Conversion of upper-limb inertial measurement unit data to joint angles: A systematic review}}.
\newblock {\emph{\JournalTitle{Sensors}}} \textbf{\bibinfo{volume}{23}} (\bibinfo{year}{2023}).

\bibitem{Cho2018}
\bibinfo{author}{Cho, Y.} \emph{et~al.}
\newblock \bibinfo{journal}{\bibinfo{title}{Evaluation of validity and reliability of inertial measurement unit-based gait analysis systems}}.
\newblock {\emph{\JournalTitle{Annals of Rehabilitation Medicine}}}  (\bibinfo{year}{2018}).

\bibitem{Kok2017}
\bibinfo{author}{Kok, M.}, \bibinfo{author}{Hol, J.~D.} \& \bibinfo{author}{Schön, T.~B.}
\newblock \emph{\bibinfo{title}{Using inertial sensors for position and orientation estimation}} (\bibinfo{publisher}{IEEE}, \bibinfo{year}{2017}).

\bibitem{W.H.K.09}
\bibinfo{author}{{de Vries}, W.}, \bibinfo{author}{Veeger, H.}, \bibinfo{author}{Baten, C.} \& \bibinfo{author}{{van der Helm}, F.}
\newblock \bibinfo{journal}{\bibinfo{title}{Magnetic distortion in motion labs, implications for validating inertial magnetic sensors}}.
\newblock {\emph{\JournalTitle{Gait \& Posture}}}  (\bibinfo{year}{2009}).

\bibitem{vanDijk2021}
\bibinfo{author}{van Dijk, M.}, \bibinfo{author}{Kok, M.}, \bibinfo{author}{Berger, M.}, \bibinfo{author}{Hoozemans, M.} \& \bibinfo{author}{Veeger, D.}
\newblock \bibinfo{journal}{\bibinfo{title}{Machine learning to improve orientation estimation in sports situations challenging for inertial sensor use}}.
\newblock {\emph{\JournalTitle{Frontiers in Sports and Active Living}}}  (\bibinfo{year}{2021}).

\bibitem{Kok2019}
\bibinfo{author}{Kok, M.} \& \bibinfo{author}{Schön, T.~B.}
\newblock \bibinfo{journal}{\bibinfo{title}{A fast and robust algorithm for orientation estimation using inertial sensors}}.
\newblock {\emph{\JournalTitle{IEEE Signal Processing Letters}}}  (\bibinfo{year}{2019}).

\bibitem{AlBorno2022}
\bibinfo{author}{Al~Borno, M.} \emph{et~al.}
\newblock \bibinfo{journal}{\bibinfo{title}{Opensense: An open-source toolbox for inertial-measurement-unit-based measurement of lower extremity kinematics over long durations}}.
\newblock {\emph{\JournalTitle{Journal of NeuroEngineering and Rehabilitation}}}  (\bibinfo{year}{2022}).

\bibitem{Weygers2020}
\bibinfo{author}{Weygers, I.} \emph{et~al.}
\newblock \bibinfo{journal}{\bibinfo{title}{{Drift-Free} {Inertial} {Sensor-Based} {Joint} {Kinematics} {for} {Long-Term} {Arbitrary} {Movements}}}.
\newblock {\emph{\JournalTitle{IEEE Sensors Journal}}}  (\bibinfo{year}{2020}).

\bibitem{winter1990biomechanics}
\bibinfo{author}{Winter, D.~A.}
\newblock \emph{\bibinfo{title}{Biomechanics and Motor Control of Human Movement}} (\bibinfo{publisher}{John Wiley \& Sons}, \bibinfo{year}{1990}), \bibinfo{edition}{2nd} edn.

\bibitem{Kuo98}
\bibinfo{author}{Kuo, A.~D.}
\newblock \bibinfo{journal}{\bibinfo{title}{A least-squares estimation approach to improving the precision of inverse dynamics computations}}.
\newblock {\emph{\JournalTitle{Journal of Biomechanical Engineering}}} \textbf{\bibinfo{volume}{120}} (\bibinfo{year}{1998}).

\bibitem{RIEMER20081503}
\bibinfo{author}{Riemer, R.} \& \bibinfo{author}{Hsiao-Wecksler, E.~T.}
\newblock \bibinfo{journal}{\bibinfo{title}{Improving joint torque calculations: Optimization-based inverse dynamics to reduce the effect of motion errors}}.
\newblock {\emph{\JournalTitle{Journal of Biomechanics}}}  (\bibinfo{year}{2008}).

\bibitem{Faber2018}
\bibinfo{author}{Faber, H.}, \bibinfo{author}{van Soest, A.~J.} \& \bibinfo{author}{Kistemaker, D.~A.}
\newblock \bibinfo{journal}{\bibinfo{title}{Inverse dynamics of mechanical multibody systems: An improved algorithm that ensures consistency between kinematics and external forces}}.
\newblock {\emph{\JournalTitle{PLoS One}}}  (\bibinfo{year}{2018}).

\bibitem{Ojeda2016}
\bibinfo{author}{Ojeda, J.}, \bibinfo{author}{Martínez-Reina, J.} \& \bibinfo{author}{Mayo, J.}
\newblock \bibinfo{journal}{\bibinfo{title}{The effect of kinematic constraints in the inverse dynamics problem in biomechanics}}.
\newblock {\emph{\JournalTitle{Multibody System Dynamics}}}  (\bibinfo{year}{2016}).

\bibitem{Dorschky2019}
\bibinfo{author}{Dorschky, E.}, \bibinfo{author}{Nitschke, M.}, \bibinfo{author}{Seifer, A.~K.}, \bibinfo{author}{van~den Bogert, A.~J.} \& \bibinfo{author}{Eskofier, B.~M.}
\newblock \bibinfo{journal}{\bibinfo{title}{{Estimation of gait kinematics and kinetics from inertial sensor data using optimal control of musculoskeletal models}}}.
\newblock {\emph{\JournalTitle{Journal of Biomechanics}}} \textbf{\bibinfo{volume}{95}} (\bibinfo{year}{2019}).

\bibitem{Haraguchi2024}
\bibinfo{author}{Haraguchi, N.} \& \bibinfo{author}{Hase, K.}
\newblock \bibinfo{journal}{\bibinfo{title}{Prediction of ground reaction forces and moments and joint kinematics and kinetics by inertial measurement units using 3d forward dynamics model}}.
\newblock {\emph{\JournalTitle{Journal of Biomechanical Science and Engineering}}}  (\bibinfo{year}{2024}).

\bibitem{Nitschke2024}
\bibinfo{author}{Nitschke, M.}, \bibinfo{author}{Dorschky, E.}, \bibinfo{author}{Leyendecker, S.}, \bibinfo{author}{Eskofier, B.~M.} \& \bibinfo{author}{Koelewijn, A.~D.}
\newblock \bibinfo{journal}{\bibinfo{title}{Estimating {3D} kinematics and kinetics from virtual inertial sensor data through musculoskeletal movement simulations}}.
\newblock {\emph{\JournalTitle{Frontiers in Bioengineering and Biotechnology}}}  (\bibinfo{year}{2024}).

\bibitem{Delp2007}
\bibinfo{author}{Delp, S.~L.} \emph{et~al.}
\newblock \bibinfo{journal}{\bibinfo{title}{Opensim: open-source software to create and analyze dynamic simulations of movement}}.
\newblock {\emph{\JournalTitle{IEEE Transactions on Biomedical Engineering}}}  (\bibinfo{year}{2007}).

\bibitem{Seth2018OpenSim}
\bibinfo{author}{Seth, A.} \emph{et~al.}
\newblock \bibinfo{journal}{\bibinfo{title}{Opensim: Simulating musculoskeletal dynamics and neuromuscular control to study human and animal movement}}.
\newblock {\emph{\JournalTitle{PLoS Comput Biol}}}  (\bibinfo{year}{2018}).

\bibitem{Seth2011}
\bibinfo{author}{Seth, A.}, \bibinfo{author}{Sherman, M.}, \bibinfo{author}{Reinbolt, J.~A.} \& \bibinfo{author}{Delp, S.~L.}
\newblock \bibinfo{journal}{\bibinfo{title}{Opensim: a musculoskeletal modeling and simulation framework for in silico investigations and exchange}}.
\newblock {\emph{\JournalTitle{Procedia IUTAM}}} \textbf{\bibinfo{volume}{2}} (\bibinfo{year}{2011}).

\bibitem{Vivian2016}
\bibinfo{author}{Vivian, M.}, \bibinfo{author}{Tagliapietra, L.}, \bibinfo{author}{Sartori, M.} \& \bibinfo{author}{Reggiani, M.}
\newblock \bibinfo{title}{Dynamic simulation of robotic devices using the biomechanical simulator opensim}.
\newblock In \emph{\bibinfo{booktitle}{Proceedings of the 13th International Conference IAS-13}} (\bibinfo{year}{2016}).

\bibitem{Kok2014}
\bibinfo{author}{Kok, M.}, \bibinfo{author}{Hol, J.~D.} \& \bibinfo{author}{Sch{\"{o}}n, T.~B.}
\newblock \bibinfo{title}{{An optimization-based approach to human body motion capture using inertial sensors}}.
\newblock In \emph{\bibinfo{booktitle}{Proceedings of the 19th World Congress The International Federation of Automatic Control}} (\bibinfo{year}{2014}).

\bibitem{Escalona21}
\bibinfo{author}{Escalona, J.~L.}, \bibinfo{author}{Urda, P.} \& \bibinfo{author}{Muñoz, S.}
\newblock \bibinfo{journal}{\bibinfo{title}{A {T}rack {G}eometry {M}easuring {S}ystem {B}ased on {M}ultibody {K}inematics, {I}nertial {S}ensors and {C}omputer {V}ision}}.
\newblock {\emph{\JournalTitle{Sensors}}}  (\bibinfo{year}{2021}).

\bibitem{Lugris2024}
\bibinfo{author}{Lugrís, U.}, \bibinfo{author}{Pérez-Soto, M.}, \bibinfo{author}{Michaud, F.} \& \bibinfo{author}{Cuadrado, J.}
\newblock \bibinfo{journal}{\bibinfo{title}{Human motion capture, reconstruction, and musculoskeletal analysis in real time}}.
\newblock {\emph{\JournalTitle{Multibody System Dynamics}}}  (\bibinfo{year}{2024}).

\bibitem{Featherstone2008}
\bibinfo{author}{Featherstone, R.}
\newblock \emph{\bibinfo{title}{Rigid {B}ody {D}ynamics {A}lgorithms}} (\bibinfo{publisher}{Springer}, \bibinfo{address}{New York}, \bibinfo{year}{2008}).

\bibitem{seth2010minimal}
\bibinfo{author}{Seth, A.}, \bibinfo{author}{Sherman, M.}, \bibinfo{author}{Eastman, P.} \& \bibinfo{author}{Delp, S.}
\newblock \bibinfo{journal}{\bibinfo{title}{Minimal formulation of joint motion for biomechanisms}}.
\newblock {\emph{\JournalTitle{Nonlinear Dynamics}}}  (\bibinfo{year}{2010}).

\bibitem{SHERMAN2011241}
\bibinfo{author}{Sherman, M.~A.}, \bibinfo{author}{Seth, A.} \& \bibinfo{author}{Delp, S.~L.}
\newblock \bibinfo{journal}{\bibinfo{title}{Simbody: multibody dynamics for biomedical research}}.
\newblock {\emph{\JournalTitle{Procedia IUTAM}}}  (\bibinfo{year}{2011}).

\bibitem{NIANDONG2017388}
\bibinfo{author}{Nian-dong, L.}, \bibinfo{author}{Kun, D.}, \bibinfo{author}{Jia-peng, T.} \& \bibinfo{author}{Wei-xin, D.}
\newblock \bibinfo{journal}{\bibinfo{title}{Analytical {S}olution of {J}acobian {M}atrices of {WDS} {M}odels}}.
\newblock {\emph{\JournalTitle{Procedia Engineering}}}  (\bibinfo{year}{2017}).
\newblock \bibinfo{note}{XVIII International Conference on Water Distribution Systems, WDSA2016}.

\bibitem{Havlík_2015}
\bibinfo{author}{Havlík, J.} \& \bibinfo{author}{Straka, O.}
\newblock \bibinfo{journal}{\bibinfo{title}{Performance evaluation of iterated extended {K}alman filter with variable step-length}}.
\newblock {\emph{\JournalTitle{Journal of Physics: Conference Series}}}  (\bibinfo{year}{2015}).

\bibitem{Morrow2017}
\bibinfo{author}{Morrow, M. M.~B.}, \bibinfo{author}{Lowndes, B.}, \bibinfo{author}{Fortune, E.}, \bibinfo{author}{Kaufman, K.~R.} \& \bibinfo{author}{Hallbeck, M.~S.}
\newblock \bibinfo{journal}{\bibinfo{title}{Validation of inertial measurement units for upper body kinematics}}.
\newblock {\emph{\JournalTitle{Journal of Applied Biomechanics}}}  (\bibinfo{year}{2017}).

\bibitem{Fang2023}
\bibinfo{author}{Fang, Z.}, \bibinfo{author}{Woodford, S.}, \bibinfo{author}{Senanayake, D.} \& \bibinfo{author}{Ackland, D.}
\newblock \bibinfo{journal}{\bibinfo{title}{Conversion of upper-limb inertial measurement unit data to joint angles: A systematic review}}.
\newblock {\emph{\JournalTitle{Sensors (Basel)}}}  (\bibinfo{year}{2023}).

\bibitem{kuka_iiwa_torque}
\bibinfo{author}{AG, K.}
\newblock \emph{\bibinfo{title}{KUKA LBR iiwa Technical Data}} (\bibinfo{year}{2015}).

\bibitem{kuka_iiwa_accuracy}
\bibinfo{author}{AG, K.}
\newblock \emph{\bibinfo{title}{KUKA LBR iiwa Specifications - Accuracy Information}} (\bibinfo{year}{2024}).

\end{thebibliography}









\end{document}